%% file: main.tex
\documentclass[acmsmall]{acmart}

\AtBeginDocument{%
  \providecommand\BibTeX{{%
    \normalfont B\kern-0.5em{\scshape i\kern-0.25em b}\kern-0.8em\TeX}}}

\setcopyright{acmcopyright}
\copyrightyear{2020}
\acmYear{2020}
\acmDOI{---}

\acmVolume{1}
\acmNumber{1}
\acmMonth{9}

\usepackage{lipsum}
\usepackage{hyperref}
\usepackage{rotating}
\usepackage{tablefootnote}
\usepackage{threeparttable}
\usepackage{commath}
\usepackage{adjustbox}
\usepackage{enumitem}
\usepackage{multirow}
\usepackage{graphicx}
\usepackage{booktabs,tabulary}
\usepackage{amssymb}
\usepackage{amsmath}
\usepackage{lipsum}
\usepackage[utf8]{inputenc}
\newcommand{\R}{\mathbb{R}}
\def\SB#1{\textsubscript{#1}}
\def\SPSB#1#2{\rlap{\textsuperscript{#1}}\SB{#2}}

\begin{document}

\title{Adversarial Machine Learning in {Image Classification}: A Survey Towards the Defender's Perspective}

\author{Gabriel R. Machado}
\email{gabriel.rmachado@ime.eb.br}
\affiliation{%
  \institution{Military Institute of Engineering (IME)}
  \city{Rio de Janeiro}
  \country{Brazil}
}
\author{Eug\^{e}nio Silva}
\affiliation{%
  \institution{State University of West Zone (UEZO)}
  \city{Rio de Janeiro}
  \country{Brazil}}
\email{eugeniosilva@uezo.rj.gov.br}
\author{Ronaldo R. Goldschmidt}
\email{ronaldo.rgold@ime.eb.br}
\affiliation{%
  \institution{Military Institute of Engineering (IME)}
  \city{Rio de Janeiro}
  \country{Brazil}
}
\renewcommand{\shortauthors}{Machado, et al.}

\begin{abstract}
    Deep Learning algorithms have achieved the state-of-the-art performance for Image Classification and have been used even in security-critical applications, such as biometric recognition systems and self-driving cars. However, recent works have shown those algorithms, which can even surpass the human capabilities, are vulnerable to adversarial examples. In Computer Vision, adversarial examples are images containing subtle perturbations generated by malicious optimization algorithms in order to fool classifiers. As an attempt to mitigate these vulnerabilities, numerous countermeasures have been constantly proposed in literature. Nevertheless, devising an efficient defense mechanism has proven to be a difficult task, since many approaches have already shown to be ineffective to adaptive attackers. Thus, this self-containing paper aims to provide all readerships with a review of the latest research progress on Adversarial Machine Learning in Image Classification, however with a defender's perspective. Here, novel taxonomies for categorizing adversarial attacks and defenses are introduced and discussions about the existence of adversarial examples are provided. Further, in contrast to exisiting surveys, it is also given relevant guidance that should be taken into consideration by researchers when devising and evaluating defenses. Finally, based on the reviewed literature, it is discussed some promising paths for future research.
\end{abstract}



\maketitle

\section{Introduction}
\input{sections/sec1.tex}

\section{Background}\label{sec:2}

\input{sections/sec2.tex}

\section{Adversarial Images and Attacks}\label{sec:3}
\input{sections/sec3.tex}

\section{Defenses against Adversarial Attacks}\label{sec:4}
\input{sections/sec4.tex}

\section{Explanations for the Existence of Adversarial Examples}\label{sec:5}
\input{sections/sec5.tex}

\section{Principles for Designing and Evaluating Defenses}\label{sec:6}
\input{sections/sec6.tex}

\section{Directions of Future Work}\label{sec:7}
\input{sections/sec7.tex}

\section{Final Considerations}\label{sec:8}
\input{sections/sec8.tex}

\bibliographystyle{ACM-Reference-Format}
\bibliography{acmart}

\appendix
\input{sections/appendix.tex}

\end{document}

%% file: sections/sec1.tex
In the last years, Deep Learning algorithms have made an important and rapid progress in solving numerous tasks involving complex analysis of raw data. Among some relevant cases, it can be mentioned major advances in speech recognition and natural language processing \cite{Dahl2012, xu2019nadaq}, games \cite{Silver1140}, finacial market analysis \cite{heaton2017deep}, fraud and malware detection \cite{Dahl2013, knorr2015}, prevention of DDoS attacks \cite{Yuan2017} and Computer Vision \cite{Krizhevsky2012, SzegedyGoogLenet2015, Szegedy2015, He2016, hu2017squeeze}. In the field of Computer Vision, the Convolutional Neural Networks (CNNs) have become the state-of-the-art Deep Learning algorithms since \citeauthor{Krizhevsky2012} \cite{Krizhevsky2012} have presented innovative results in image classification tasks using the AlexNet architecture. Thereafter, motivated by the continuous popularization of GPUs and frameworks, the CNNs have kept growing in performance, being currently even used in security-critical apllications, such as medical sciences and diagnostics \cite{cheng2016computer, litjens2017survey}, autonomous vehicles \cite{bojarski2016end}, surveillance systems \cite{ding2018deep} and biometric and handwritten characterers recognition \cite{bae2018secure, tolosana2018exploring, srivastava2019optical}.

{Nevertheless, some researchers have begun to argue if {the same deep learning algorithms, which could even surpass the human performance \cite{karpathy2014learned}}, were actually robust enough to be used in safety-critical environments. Unfortunately, since the paper of \citeauthor{Szegedy2013} \cite{Szegedy2013}, various works have highlighted the vulnerability of deep learning models in different tasks such as speech recognition \cite{audio_carlini}, text classification \cite{ebrahimi2017hotflip}, malware detection \cite{grosse_malware} and specially image classification \cite{papernot2016limitations, Goodfellow2014, carlini2017towards} before \textit{adversarial attacks}. Adversarial attacks are usually conducted in the form of subtle perturbations generated by an optimization algorithm and inserted into a legitimate image in order to produce an adversarial example which, in the field of Computer Vision is specifically known as \textit{adversarial image}. After being sent to be classified, an adversarial image is often able to lead CNNs to produce a prediction different from the expected, usually with a high confidence. Adversarial attacks on image classifiers are the most common in the literature and, for this reason, are the focus of this paper.}

The vulnerability of CNNs and other Deep Learning algorithms to adversarial attacks have forced the scientific community to revisit all the processes related to the construction of intelligent models, from the elaboration of architectures to the formulation of the training algorithms used, as a attempt to hypothesize some possible reasons concerning this lack of robustness\footnote{Robustness can be defined as the capacity of a model or defense to tolerate adversarial disturbances by delivering reliable and stable outputs \cite{Xiao2017}.} and thus propose countermeasures that may hold future attacks of adversarial nature. This arms race between attacks and defenses against adversarial examples has ended up forming a recent research area called \textit{Adversarial Machine Learning} that, in a nutshell, struggles to construct more robust Deep Learning models. 

{Adversarial Machine Learning in Image Classification} is currently a very active research path which is responsible for most of the work in the area, with novel papers produced almost daily. However, there is neither a known efficient solution for securing Deep Leanirng models nor any fully accepted explanations for the existence of adversarial images yet. Taking into account the dynamism and relevance of this research area, it is crucial to be available in literature comprehensive and up-to-date review papers in order to position and orientate their readers about the actual scenario. {Although there are already some extensive surveys \cite{akhtar2018threat, yuan2019adversarial, wiyatno2019adversarial}, they have already become somewhat outdated due to the great activity in the area. Moreover, they bring out a general overview of the Adversarial Machine Learning field, what, in turn, contributes to these papers neither have focused enough in works that have proposed defenses against adversarial attacks nor have provided proper guidance for those who wishes to invest in novel countermeasures.}

Therefore, keeping in mind the importance of Adversarial Machine Learning in Image Classification to the development of more robust defenses and architectures against adversarial attacks, this self-contained paper aims to provide for all readerships an exhaustive and detailed review of the {literature, however with a defender's perspective. The present survey covers from the background needed to clarify the reader essential concepts in the area to the techinical formalisms related to adversarial examples and attacks. Futhermore, a comprehensive survey of defenses and coutermeasures against adversarial attacks is made and categorized on a novel taxonomy}. Then, based on the works of \citeauthor{carlini2017bypassing} \cite{carlini2017bypassing} and \citeauthor{carlini2019evaluating} \cite{carlini2019evaluating}, {the present paper discusses} some principles for designing and evaluating defenses which are intended to guide researchers to introduce stronger security methods. Essentially, the main contributions of this work are the following:

\begin{itemize}[leftmargin=*]
    \item The update of some existing taxonomies in order to categorize different types of adversarial images and novel attack approaches that have raised in literature;\vspace{2mm}
    
    \item The discussion and organization of defenses against adversarial attacks based on a novel taxonomy;\vspace{2mm}
    
    \item The address of relevant explanations for the existence of adversarial examples;\vspace{2mm}
    
    \item {The provision of some important guidances that should be followed by researchers when devising and evaluating defenses;}\vspace{2mm}
    
    \item {The discussion of promising research paths for future works in the field.}\vspace{2mm}
\end{itemize}

The remaining of this paper is structured as follows: Section \ref{sec:2} brings an essential background which covers important topics and concepts for the proper understanding of this work. Section \ref{sec:3} formalizes and also categorizes adversarial examples and attacks. {Section \ref{sec:4} makes a deep review on defenses existing in literature and proposes a novel taxonomy for organizing them.} Section \ref{sec:5} addresses and formalizes relevant explanations for the existence of adversarial examples that have supported the development of attacks and defenses. Section \ref{sec:6} provides close guidance based on relevant work to help defenders and reviewers to respectivelly design and evaluate security methods. Section \ref{sec:7} lists promising research paths in Adversarial Machine Learning for Image Classification. Finally, Section \ref{sec:8} brings the final considerations.

%% file: sections/sec2.tex
Conventional Machine Learning models (also known as \textit{shallow models} \cite{bengio2007scaling}) have begun to have high dependency of domain experts and present critical limitations when attempting to extract useful patterns from complex data, such as images and audio speech \cite{yuan2019adversarial}. Therefore, it has been necessary to develop traditional learning algorithms into more elaborated architectures, forming a recent area in Artificial Intelligence called \textit{Deep Learning} \cite{hinton2007learning}.
Deep Learning is a subfield of Machine Learning where its algorithms simulate the operation of the human brain in order to extract and learn hidden representations from raw inputs, oftentimes without any human intervention. Mostly, Deep Learning is based on Deep Neural Networks (DNNs), which are formed by many layers containing numerous processing units, which gathers knowledge from a massive amount of data by appling several linear and non-linear transformations on the received inputs, which in turn, allows these models to learn high-level abstractions from simpler concepts \cite{sambit2018, Goodfellow-et-al-2016, Guo2016}.

This paper focuses mostly on Convolutional Neural Networks. CNNs are a special type of Deep Neural Network and currently are the state-of-the-art algorithms for Image Classification \cite{vorobeychik2018adversarial}. However, {Appendix A briefly covers other tasks Adversarial Machine Learning takes part in Computer Vision. The next section explains, in a nutshell, the main components of a CNN, in addition to list the state-of-the-art architectures over the years, according to the ILSVRC challenge \cite{imagenetRussakovsky}.}

\subsection{Convolutional Neural Networks}

\begin{figure}[h!]
	\centering
	\includegraphics[width=4in]{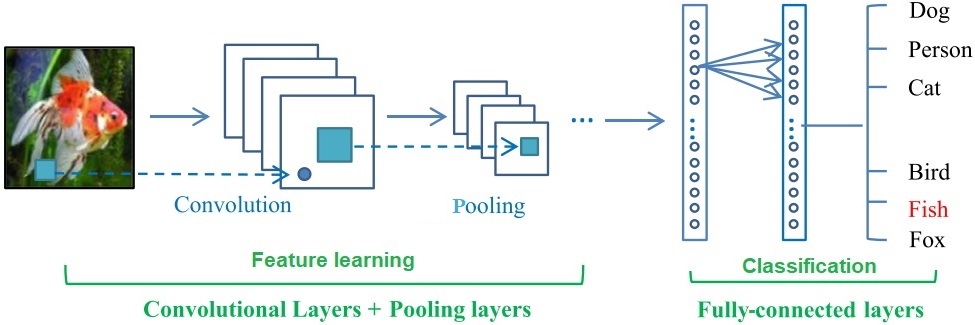}
	\caption{The standard architecture of a CNN. Adapted from \citeauthor{Guo2016} \cite{Guo2016}.}
	\label{fig:cnn_pipeline}
\end{figure}

CNN architectures usually perform feature learning by making use of \textit{(i) convolution} and \textit{(ii) pooling} layers which, respectivelly, extracts useful features from images and reduces their spatial dimensionality. After feature learning, comes the fully connected layer (FC), which works in a way similar to a common neural network. In a classification task, FCs produce a single probability vector as output, which is called the \textit{probability vector}. The probability vector contains membership probabilities of a given input $x$ corresponding to each class $c_i \in C_n$, where $C_n$ is the set containing all the $n$ classes belonging to the original problem. Summing up all the probabilities must result in 1, and the chosen class for $x$ is the one which has the highest membership probability. Figure \ref{fig:cnn_pipeline} depicts an example of a standard architecture of a CNN.

\begin{figure}[h!]
    \centering
    \includegraphics[scale=0.325]{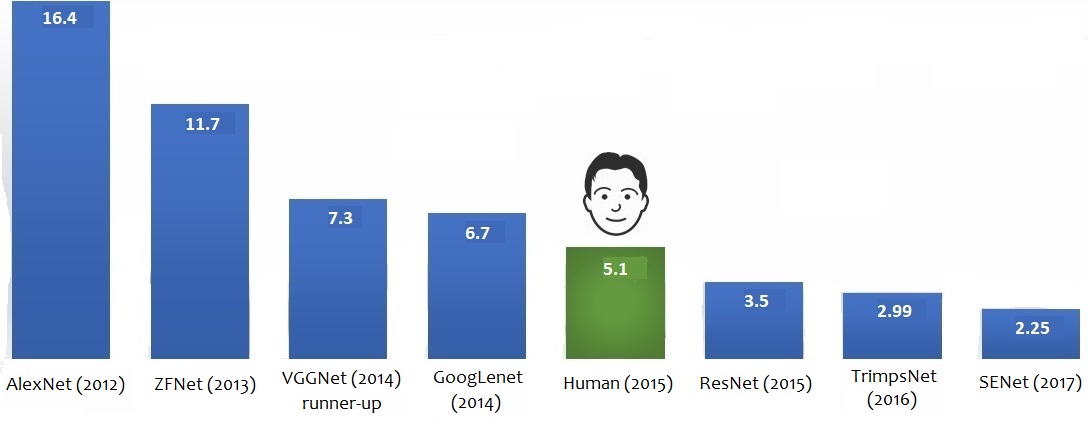}
    \caption[Performance of relevant CNN architectures in the ILSVRC top-5 error rate classification challenge over the years \cite{imagenetRussakovsky}. Since 2015, the CNNs have surpassed the human's performance \cite{karpathy2014learned}.]{Top-5 error rate\footnotemark of winning CNN architectures in ILSVRC classification challenge over the years \cite{imagenetRussakovsky}. Since 2015, the CNNs have surpassed the human's performance \cite{karpathy2014learned}.}
    \label{fig:imagenet_contest}
    \vspace{-5mm}
\end{figure}

An important contest in Computer Vision, called ILSVRC (ImageNet Large Scale Visual Recognition Challenge) \cite{imagenetRussakovsky}, has encouraged until 2017 the creation of more accurate CNN architectures. Figure \ref{fig:imagenet_contest} shows some relevant CNN architectures over the years in ILSVRC challenge, namely AlexNet \cite{Krizhevsky2012}, ZFNet \cite{zeiler2014visualizing}, VGGNet \cite{vgg2014}, GoogLeNet \cite{SzegedyGoogLenet2015}, ResNet \cite{resnet2016}, TrimpsNet\footnote{There has been no novel scientific contribution which justified the production of a paper, and for this reason, the authors of TrimpsNet only shared the results using the ImageNet and COCO joint workshop in ECCV 2016, which are available at: \url{http://image-net.org/challenges/talks/2016/Trimps-Soushen@ILSVRC2016.pdf}. Accessed in February 12, 2020.} and SENet \cite{hu2017squeeze}. Since 2015, CNNs have surpassed the human performance \cite{karpathy2014learned}.

\footnotetext{In contrast to the traditional top-1 classification, a top-5 classification considers a
model had a correct predition when, given a test sample, its true class is among the five highest output probabilites predicted by this model.}

\subsection{Other Deep Learning Algorithms}\label{sec:other_deep}

Apart from CNNs, there are other important Deep Learning architectures which are frequently used in Adversarial Machine Learning, such as Autoencoders (AEs) and Generative Adversarial Networks (GANs). The next sections describe these architectures in more details. 

\subsubsection{Autoencoders}

\begin{figure}[h!]
	\centering
	\includegraphics[width=3.5in]{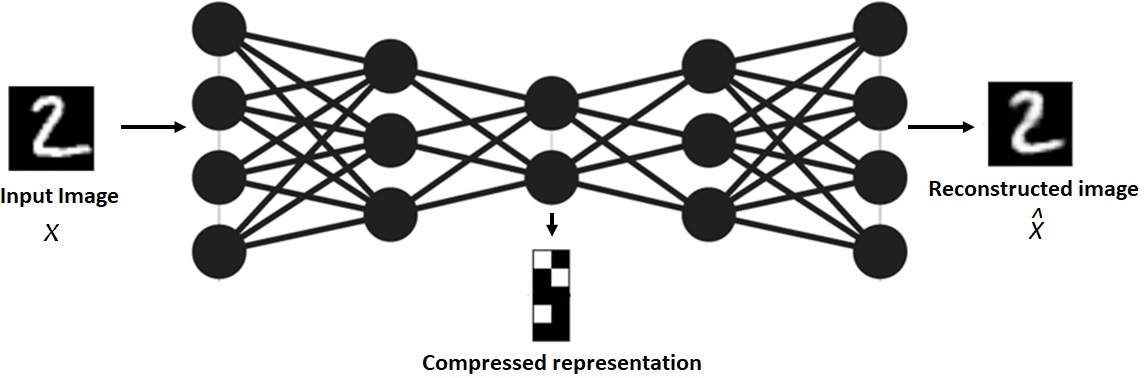}
	\caption{{An example of an autoencoder architecture.}}
	\label{fig:autoencoder}
	\vspace{-3mm}
\end{figure}

An autoencoder is a neural network which aims to approximate its output to an input sample, or in other words, it tries to approximate an input $x$ to its \textit{identity function} by generating an output $\hat{x}$ as similiar as possible to $x$ from a compressed representation learnt. {An example of an autoencoder architecture is depicted by Figure \ref{fig:autoencoder}.} Despite looking a trivial task, the autoencoder is actually trying to learn the inner representations of the input, regarding the structure of data. Autoencoders are useful for two main purposes: (i) dimensionality reduction, retaining only the most important data features \cite{hinton2006reducing, Liu2017a, schmidhuber2015deep} and (ii) data generation process \cite{deng2012three}. 


\subsubsection{Generative Adversarial Networks}

Generative Adversarial Networks (GANs) are a framework introduced by Goodfellow et al. \cite{goodfellow2014generative} for building \textit{generative models} $\mathcal{P}_{G}$ which resembles the data distribution $\mathcal{P}_{data}$ used in the training set. GANs can be used to improve the representation of data, to conduct \textit{unsurpevised learning} and to even construct defenses against adversarial images \cite{yuan2019adversarial, goodfellow2014generative}. There are works that have also used GANs for other purposes, such as image-to-image translation and visual style transfer \cite{zhu2017unpaired, isola2017image}. The GANs are composed by two models (usually deep networks) trained simultaneously: a \textit{generator} $G$ and a \textit{discriminator} $D$. The \textit{generator} receives an input $x$ and tries to generate an output $z$ from a probability distribution $\mathcal{P}_G$. In contrast, the \textit{discriminator} classifies $z$, producing a label that determines if $z$ belongs to the distribution $\mathcal{P}_{data}$ (\textit{benign} or \textit{real input}) or $\mathcal{P}_G$ (\textit{fake} or \textit{adversarial input}). In other words, the \textit{generator} $G$ is actually being trained to fool the classifier $D$. In this competing scenario, GANs are usually capable of generating data samples that looks close to benign examples.

%% file: sections/sec3.tex
\begin{figure}[h!]
    	\centering
    	\includegraphics[width=0.8\textwidth]{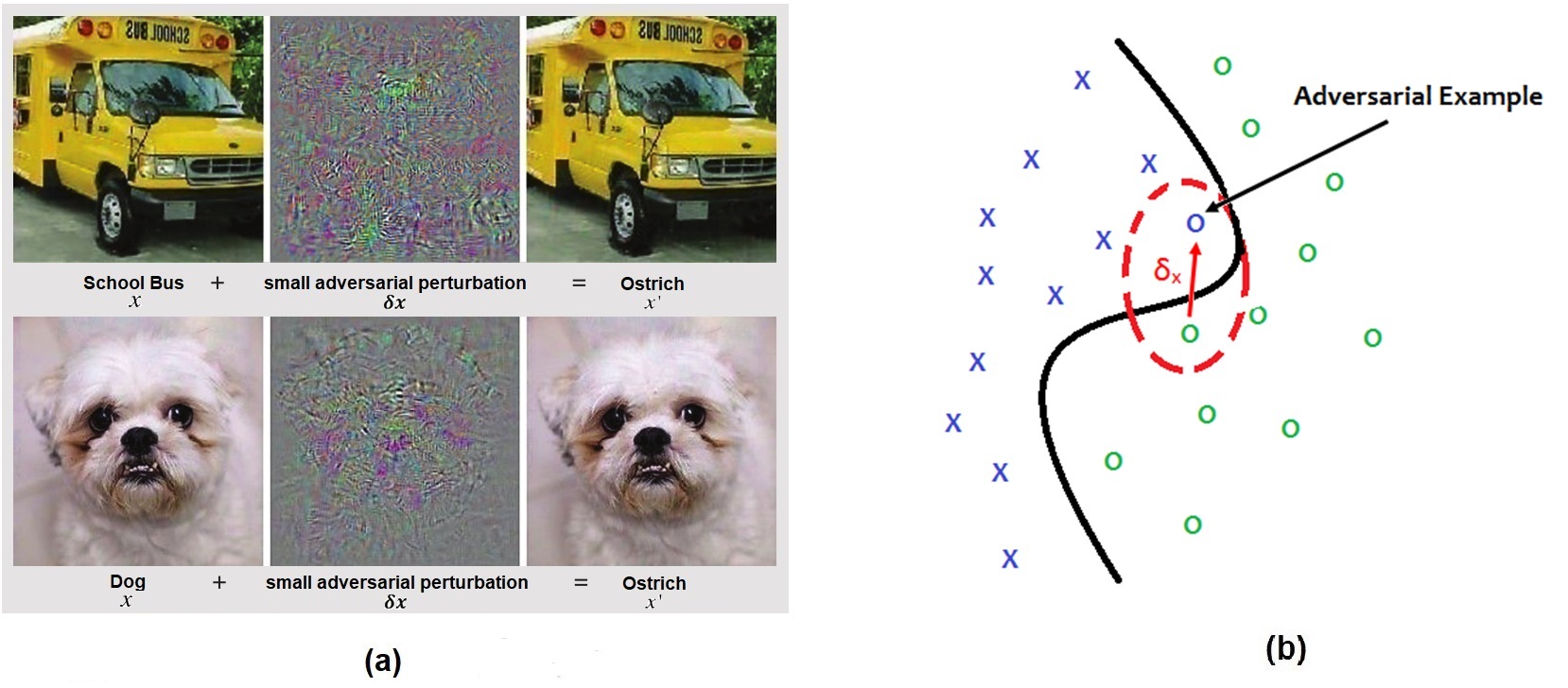}
    	\caption{(a): Malicious and usually imperceptible perturbations present in a input image can induce trained models to misclassification. Adapted from \citeauthor{klarreich2016learning} \cite{klarreich2016learning}. (b): The objective of an adversarial attack is to generate a perturbation $\delta x$ and insert it into a legitimate image $x$ in order to make the resulting adversarial image $x' = x + \delta x$ cross the decision boundary. Adapted from \citeauthor{bakhti2019ddsa} \cite{bakhti2019ddsa}.}
    	\label{fig:dog_adv}
    	\vspace{-5mm}
    \end{figure}

{Formally, an adversarial image can be defined as follows:} let $f$ be a classification model trained with legitimate images (\textit{i.e.} images which do not have any malicious perturbations) and let $x$ be a legitimate image (where $x \in \R^{w \times h \times c}$, such that $w$ and $h$ are the dimensions of the image and $c$ is its amount of color channels), then it is crafted, from $x$, an image $x'$, such that $x' = x + \delta x$, where $\delta x$ is the perturbation needed to make $x$ cross the decision boundary, resulting $f(x) \neq f(x')$ {(see Figure \ref{fig:dog_adv}a)}. The perturbation $\delta x$ can also be interpreted as a vector $\vec{\delta x}$, where its magnitude $\|\vec{\delta x}\|$ represents the amount of perturbation needed to translate the point represented by the image $x$ in the space beyond the decision boundary. {Figure \ref{fig:dog_adv}b illustrates a didactic example of inserting a perturbation $\delta x$ into a legitimate image $x$ on a 2D space.} According to \citeauthor{Cao} \cite{Cao}, an adversarial image is considered optimal if it satisfies two requirements: (i) if the perturbations inserted on this image are imperceptible to human eyes and (ii) if these perturbations are able to induce the classification model to produce an incorrect output, preferably with a high confidence.
    
\subsection{Taxonomy of Adversarial Images} \label{sec:tax_advimages}

This section is based on the works of \citeauthor{Barreno2010, huang2011adversarial, yuan2019adversarial, Kumar2017, Xiao2017} and \citeauthor{ brendel2017decision} \cite{Barreno2010, huang2011adversarial, yuan2019adversarial, Kumar2017, Xiao2017, brendel2017decision} to propose a broader\footnote{{For comparative purposes, the novel topics proposed by this paper are underlined.}} taxomomy to adversarial images formed by three diferent axes: \textit{(i) perturbation scope}, \textit{(ii) perturbation visibility} and \textit{(iii) perturbation measurement}. The next sections explain each axis in details.

\subsubsection{Perturbation Scope}

Adversarial images may contain individual-scoped perturbations or universal-scoped perturbations.

\begin{itemize}[leftmargin=*]
    \item \textbf{Individual-scoped perturbations:} individual-scoped perturbations are the most common in literature. They are generated individually for each input image;
    
    \item \textbf{Universal-scoped perturbations:} universal-scoped perturbations are image-agnostic perturbations, \textit{i.e.} they are perturbations generated independently from any input sample. Nevertheless, when they are applied to an legitimate image, the resulting adversarial example is often able to lead models to misclassification \cite{metzen2017universal, Moosavi-Dezfooli2017}. Universal perturbations permit adversarial attacks being conducted more easily in real-word scenarios, since these perturbations are crafted just once to be inserted into any sample belonging to a certain dataset. 
\end{itemize}

\subsubsection{Perturbation Visibility}

The efficiency and visibility of perturbations can be organized as:

\begin{itemize}[leftmargin=*]
    \item \textbf{Optimal perturbations:} these perturbations are imperceptible to human eyes, but are useful to lead deep learning models to misclassification, usually with a high confidence on the prediction;
    
    \item \textbf{\underline{Indistinguishable perturbations:}} indistinguishable perturbations are also imperceptible to human eyes, however they are insufficient to fool deep learning models; 
    
    \item \textbf{\underline{Visible perturbations:}} perturbations that, when inserted into a image, are able to fool deep learning models. However they can also be easily spotted by humans \cite{karmon2018lavan, brown2017adversarial};
    
    \item {\textbf{\underline{Physical perturbations:}} are perturbations designed outside the digital scope and physically added to real-world objects themselves \cite{Evtimov2017}. Although some works have adapted physical perturbations to Image Classification \cite{Kurakin2016a}, they are usually directed to tasks involving Object Detection \cite{Evtimov2017, chen2018shapeshifter, thys2019fooling} (see Appendix C);}
    
    \item {\underline{\textbf{Fooling images:}}} perturbations which corrupt images to the point of making them unrecognizable by humans. Nevertheless, the classification models believe these corrupted images belong to one of the classes of the original classification problem, sometimes assigning to them a high confidence on the prediction \cite{Nguyen2015}. Fooling images are also known as \textit{rubbish class examples} \cite{Goodfellow2014};
    
    \item \underline{\textbf{Noise}:} in contrast to the malicious nature of perturbations, noises are non-malicious or non-optimal corruptions that may be present or inserted into a input image. An example of noise is the gaussian noise.
\end{itemize}

\subsubsection{Perturbation Measurement}

Given the fact that it is difficult to define a metric that measures the capability of human vision, the \textit{p-norms} are most used to control the size and the amount of the perturbations that are inserted into a image \cite{Meng2017}. The p-norm $L_p$ computes the distance $\norm{x - x'}_p$ in the input space between a legitimate image $x$ and the resulting adversarial example $x'$, where $p \in \{0, 1, 2, \infty\}$. In Equation \ref{eq:2} is defined the p-norm when $p=1$ (Manhattan Distance) and $p=2$ (Euclidean Distance):
\vspace{-4mm}

\begin{equation}\label{eq:2}
    L_p = \sqrt[p]{\sum\abs{x - x'}^{p}}
\end{equation}

\noindent When $p=0$, it is counted up the number of pixels that have been modified in a legitimate sample in order to generate the adversarial image. On the other hand, the $L_\infty$ measures the maximum difference among all pixels in the corresponding positions between two images. For $L_\infty$ norm, each pixel is allowed to be modified within a maximum limit of perturbation, without having any restriction for the number of modified pixels. Formally, $L_\infty = \norm{x - x'}_{\infty} = \max\Big({\abs{x_1 - x'_1}, \abs{x_2 - x'_2}, \cdots, \abs{x_n - x'_n}}\Big)$. Despite the norms $p \in \{0, 1, 2, \infty\}$ be the most used when computing perturbations, there are some works that have defined custom metrics, as can be seen in Table \ref{tbl:attacks}.

\subsection{{Taxonomy of Attacks and Attackers}} \label{sec:tax_threat}

This section is also based on the concepts and definitions of the works of \citeauthor{Barreno2010, Kumar2017, Xiao2017, brendel2017decision, akhtar2018threat} and \citeauthor{yuan2019adversarial} \cite{Barreno2010, Kumar2017, Xiao2017, brendel2017decision, akhtar2018threat, yuan2019adversarial} {to extend\footnote{Again here, the novel topics proposed by this paper are highlighted by undelined font.} existing taxonomies which organize attacks and attackers. In the context of security, adversarial attacks and attackers are categorized under \textit{threat models}. A threat model defines the conditions under which a defense is designed to provide security garantees against certain types of attacks and attackers \cite{carlini2019evaluating}.} Basically, a threat model delimiters (i) the knowledge an attacker has about the targeted classifier (such as its parameters and architecture), (ii) his goal with the adversarial attack and (iii) how he will perform the adversarial attack. A threat model can be then classified into six different axes: \textit{(i) attacker's influency}, \textit{(ii) attacker's knowledge}, \textit{(iii) security violation}, \textit{(iv) attack specificity}, \textit{(v) attack computation} and \textit{(vi) attack approach}. 

\subsubsection{Attacker's Influence}

This axis defines how the attacker will control the learning process of deep learning models. According to \citeauthor{Xiao2017} \cite{Xiao2017}, the attacker can perform two types of attack, taking into account his influence on the classification model: (i) \textit{causative} or \textit{poisoning attacks} and (ii) \textit{evasive} or \textit{exploratory attacks}.

\begin{itemize}[leftmargin=*]
    \item \textbf{Causative or poisoning attacks}: in causative attacks, the attacker has influence on the deep learning model during its training stage. In this type of attack, the training samples are corrupted or the training set is polluted with adversarial examples in order to produce a classification model incompatible with the original data distribution; 

    \item \textbf{Evasive or exploratory attacks}: in constrast of causative attacks, in evasive attacks the attacker has influence on the deep learning models during the inference or testing stage. Evasive attacks are the most common type of attack, where the attacker craft adversarial examples that lead deep learning models to misclassification, usually with a high confidence on the prediction. Evasive attacks can also have an exploratory nature, where the attacker's objective is to gather information about the targeted model, such as its parameters, architectures, cost functions, etc. The most common exploratory attack is the \textit{input/output} attack, where the attacker provides the targeted model with adversarial images crafted by him. Afterwards, the attacker observes the outputs given by the model and tries to reproduce a substitute or surrogate model, so that it can be similar to the targeted model. The \textit{input/output} attack is usually the first step to perform black-box attacks (see Section \ref{sec:black_white_box_attacks}).
\end{itemize}

\subsubsection{Attacker's Knowledge}\label{sec:black_white_box_attacks}

Taking into consideration the attacker's knowlegde with respect to the targeted model, three types of attacks can be performed: \textit{(i) white-box attacks}, \textit{(ii) black-box attacks} and \textit{(iii) grey-box attacks}.

\begin{itemize}[leftmargin=*]
    \item \textbf{White-box attacks}: in a white-box attack, the attacker has fully access to the model's and even the defense's parameters and architectures, whenever such defense exists. This attack scenario probably would be the least frequent in real-world applications, due to the adoption of protection measures (such as users control, for example) in order to prevent unauthorized people access to the system components. By contrast, white-box attacks are usually the most powerful type of adversarial attack, and for this reason, are commonly used to evaluate the robustness of defenses and/or classification models when they are undergone to harsh conditions. Unfortunately, elaborating countermeasures resistant to white-box attacks is, so far, an open problem;
    
    \item \textbf{Black-box attacks}: in this scenario, the attacker neither has access nor knowledge about any information concerning the classification model and the defense method, when present. Black-box attacks impose more restrictions to attackers, nonetheless they are important when reproducing external adversarial attacks aiming deployed models, which in turn, better represent real-world scenarios \cite{Papernot2016c}. Despite the greater difficulty to perform black-box attacks, the attacker still might be able to evade the target model due to the \textit{transferability of adversarial examples}. Works such as \citeauthor{Szegedy2013} and \citeauthor{Papernot2016c} \cite{Szegedy2013, Papernot2016c} have shown the malicious effect of an adversarial image, generated using a certain classifier, is able to transfer and fool other classifiers, including the ones created by different learning algorithms (check Section \ref{subsec:transferability} for more details). With this property in favor of the attacker, he can create an empirical model through a causative attack called \textit{substitute} or \textit{surrogate model}, which has similar parameters to the targeted model's. Therefore, the attacker can use this surrogate model to craft adversarial images and, afterwards, deploy them to be, oftentimes, misclassified by the targeted model;
    
    \item \underline{\textbf{Grey-box attacks}:} this attack scenario has been firstly proposed by \citeauthor{Meng2017} \cite{Meng2017}. In grey-box attacks, the attacker has access to the classification model, but does not have access to any information concerning the defense method. Grey-box attacks are an intermediate alternative to evaluate defenses and classifiers, since they impose a greater threat level when compared to the black-box attacks, but without giving a wide advantage to the attacker when providing him as well with all the information concerning the defense method, as performed in white-box scenarios.
\end{itemize}

\subsubsection{Security Violation}

Security violations are often associated with the attacker's objective when performing an adversarial attack against a classifier. The security violations caused by adversarial attacks can affect the (i) \textit{integrity}, (ii) \textit{availability} and the (iii) \textit{privacy} of the targeted classifiers.

\begin{itemize}[leftmargin=*]
    \item \textbf{Integrity violation}: this is the most common violation provoked by an adversarial attack. The integrity is affected when adversarial images, crafted by an certain attacker, are able to stealthily bypass existing countermeasures and lead targeted models to misclassification, but without compromising the functionality of the system;
    
    \item \textbf{Availability violation}: occurs when the functionality of the system is also compromised, causing a denial of service. Availability violations mainly affect the reliability of learning systems by raising uncertainty of their predictions;
    
    \item \textbf{Privacy violation}: happens when the attacker is able to gain access to relevant information regarding the targeted model, such as its parameters, architecture and learning algorithms used. Privacy violations in deep learning are strictly related to black-box attacks, where the attacker queries the targeted model in order to reverse-engineer it and produce a surrogate model, which crafts adversarial examples closer to the original data distribution.
\end{itemize}

\subsubsection{Attack Specificity}

With respect to specificity, an attacker can perform a \textit{(i) targeted attack} and an \textit{(ii) untargeted (or indiscriminate) attack}. Targeted attacks aim to craft an adversarial image in order to lead the model to misclassify it in a predetermined class, chosen beforehand by the attacker. On the other hand, in untargeted attacks, the attacker just seeks to fool the model by aiming any class different from the legitimate class corresponding to the original example. Formally, let $x$ be a legitimate image, $y$ the original class of the image $x$ and $f$ a classification model; then, an adversarial image $x' = x + \delta x$ is crafted from $x$. In a targeted attack, the attacker seeks to craft a perturbation $\delta x$ in order to produce in $f$ as output a specific class $y'$, such that $f(x + \delta x) = y'$ and $y' \neq y$. Conversely, in an untargeted attack, it is generated an adversarial image $x'$, such that $f(x) \neq f(x')$. Targeted attacks usually present higher computational costs when compared to untargeted attacks. 

\subsubsection{Attack Computation}

The algorithms used to compute perturbations can be \textit{(i) sequential} and \textit{(ii) iterative}. The sequential algorithms compute in just one iteration, the perturbation that will be inserted into a legitimate image. Iterative algorithms, in turn, make use of more iterations in order to craft the perturbation. Since iterative algorithms make use of more iterations to compute perturbations, they have a higher computational cost when compared to sequential algorithms. However, the perturbations generated by iterative algorithms are usually smaller and more eficient to fool classification models than those generated by one-step procedures. 

\subsubsection{Attack Approach}

Adversarial attacks can also be organized with respect to the approach used by the attack algorithm to craft the perturbation. According to \cite{brendel2017decision}, the approach of adversarial attacks can be based on \textit{(i) gradient}, \textit{(ii) transferibility/score}, \textit{(iii) decision} and \textit{(iv) approximation}.

\begin{itemize}[leftmargin=*]
    \item \textbf{Gradient-based attacks:} this attack approach is the most used in literature. The gradient-based algorithms make use of detailed information of the target model concerning its gradient with respect to the given input. This attack approach is usually performed in white-box scenarions, when the attacker has full knowledge and access to the targeted model;
    
    \item \textbf{Transfer/Score-based attacks:}
    these attack algortihms either depend on getting access to the dataset used by the targeted model or the scores predicted by it in order to approximate a gradient. Usually, the outputs obtained by querying a targeted deep neural network are used as scores. These scores are then used along with the training dataset to fit a surrogate model which will craft the perturbations that will be inserted into the legitimate images. This attack approach is often useful in black-box attacks;
    
    \item \underline{\textbf{Decision-based attacks:}} this approach has been firstly introduced by \citeauthor{brendel2017decision} \cite{brendel2017decision}, and it is considered by the authors as a simpler and more flexible approach, since requires few changes in parameters than gradient-based attacks. A decision-based attack usually queries the softmax layer of the targeted model and, iteratively, computes smaller perturbations by using a process of rejection sampling;
    
    \item \underline{\textbf{Approximation-based attacks:}} attacks algorithms based on this approach try to approximate a gradient for some targeted model or defense formed by a non-differentiable technique usually by applying numerical methods. These approximated gradients are then used to compute adversarial perturbations.
    
\end{itemize}

\subsection{Algorithms for Generating Adversarial Images}

In Computer Vision, the algorithms used to generate adversarial perturbations are optimization methods that usually explore generalization flaws in pretrained models in order to craft and insert perturbations into legitimate images. The next sections will describe with more details four attack algorithms frequently used, namely \textit{(i) FGSM} \cite{Goodfellow2014}, \textit{(ii) BIM} \cite{Kurakin2016a}, \textit{(iii) DeepFool} \cite{Moosavi-Dezfooli2015} and \textit{(iv) CW Attack} \cite{carlini2017towards}. Afterwards, Table \ref{tbl:attacks} organizes, according to the taxonomies presented in Sections \ref{sec:tax_advimages} and \ref{sec:tax_threat}, other important attack algorithms. 

\subsubsection{Fast Gradient Sign Method (FGSM)}

FGSM is a sequential algorithm proposed by \citeauthor{Goodfellow2014} \cite{Goodfellow2014} to sustain his linear hypothesis for explaining the existence of adversarial examples (see Section \ref{subsec:linear_hyp}). The main characteristic of FGSM is its low computational cost, resulted from perturbing, in just one step (limited by a given upper bound $\epsilon$), a legitimate image at the direction of the gradient that maximizes the model error. Despite its efficiency, the perturbations generated by FGSM are usually greater and less effective to fool models than the perturbations generated by iterative algorithms. Given an image $x \in \mathbb{R}^{w \times h \times c}$, FGSM generates an adversarial image $x'$ according to Equation \ref{eq:fgsm}:

    \begin{equation}\label{eq:fgsm}
        x' = x - \epsilon \cdot sign(\vec{\nabla}_x J(\Theta, x, y))
    \end{equation}

\noindent where $\vec{\nabla}_x$ represents the gradient vector, $\Theta$ represents the network parameters, $y$ the class associated to $x$, $\epsilon$ the maximum amount of perturbation that can be inserted into $x$ and $J(\Theta, x, y)$ the cost function used to train the neural network.

\subsubsection{Basic Iterative Method (BIM)}
This attack is a iterative version of FGSM, initially proposed by \citeauthor{Kurakin2016a} \cite{Kurakin2016a}. In constrast to FGSM, BIM executes several minor steps $\alpha$, where the total size of the perturbation is limited by an upper bound defined by the attacker. Formally, BIM can be defined as a recursive method, which generates $x'$ according to Equation \ref{eq:bim}:

    \begin{equation}\label{eq:bim}
        x' = 
        \begin{cases}
            x'_0 = 0 \\
            x'_i = x'_{i-1} - \mathit{clip}(\alpha \cdot \mathit{sign}({\vec{\nabla}_x J(\Theta, x'_{i-1}, y)})
        \end{cases}
    \end{equation}

where $clip$ limits values to the lower and higher edges  outside the given interval.

\subsubsection{DeepFool}

The main ideia behind DeepFool, proposed by \citeauthor{Moosavi-Dezfooli2015} \cite{Moosavi-Dezfooli2015}, consists of finding the nearest decision boundary of a given legitimate image $x$ and then subtly perturb this image to make it cross the boundary and fool the classifier. Basically, DeepFool approximates, for each iteration, the solution of this problem by linearizing the classifier around an intermediate $x'$. The intermediate $x'$ is then updated towards the direction of an optimal direction by a small step $\alpha$. This process is repeated until the small perturbation computed by DeepFool makes $x'$ cross the decision boundary. Similarly to FGSM, DeepFool is also based on the linearity hypothesis to craft perturbations.

\subsubsection{{Carlini \& Wagner Attack}}

The CW attack has been proposed by \citeauthor{carlini2017towards} \cite{carlini2017towards} and currently represents the state-of-the-art algorithm for generating adversarial images. Formally, given an DNN $f$ having a logits layer $z$ and a input image $x$ belonging to a class $t$, CW uses the gradient descent to solve iteratively Equation \ref{eq:cw1}:

\begin{equation}\label{eq:cw1}
    \text{minimize } ||{x - x'}||\SPSB{2}{2} + c \cdot \ell(x')
\end{equation}
 where, for $x$, the attack seeks for a small perturbation $\delta_x = x - x'$ that is able to fool the classifier. To do so, a hyperparameter $c$ is used as an attempt to compute the minimal amount of perturbation required. Besides $c$, there is the cost function $\ell(x')$, which is defined according to Equation \ref{eq:cw2}.

\begin{equation}\label{eq:cw2}
    \ell(x') = \mathit{max}{(\mathit{max}\{{z(x')_i : i \neq t}\} - z(x')_t, - conf)}
\end{equation} 

In Equation \ref{eq:cw2}, the hyperparameter $conf$ refers to the attack confidence rate. Higher $conf$ values contribute to generate adversarial images capable of fooling models with a high confidence rate, \textit{i.e.} with predictions reaching probabilities up to 100\% in a incorrect class $t' \neq t$. On the other hand, higher $conf$ values also produce adversarial images usually containing larger perturbations which are easily perceptible by humans. 


\begin{table}[h!]
\centering
\begin{adjustbox}{max width=\textwidth}
\begin{threeparttable}
\caption{Main adversarial attack algorithms in Computer Vision.\tnote{*}}
 {\scriptsize 
  \setlength\tabcolsep{4pt} 
\begin{tabular}{lcccccc}
\toprule
Algorithm and Reference & Perturbation Scope & Perturbation Visibility & Perturbation Measurement & Attacker's Knowledge & Attack Specificity & Attack Approach \\ \hline
FGSM \cite{Goodfellow2014} & individual & optimal, visible & $L_\infty$ & white-box & untargeted & gradient \\
JSMA \cite{papernot2016limitations} & individual & optimal & $L_0$ & white-box & targeted & gradient \\
L-BFGS \cite{Szegedy2013} & individual & optimal & $L_\infty$ & white-box & targeted & gradient \\ POBA-GA \cite{pobga2019} & individual & optimal & custom & black-box & targeted, untargeted & decision \\ AutoZoom \cite{tu2019autozoom} & individual & optimal & $L_2$ & black-box & targeted, untargeted & decision \\
DeepFool \cite{Moosavi-Dezfooli2015} & individual, universal & optimal & $L_1$, $L_2$, $L_\infty$ & white-box & untargeted & gradient \\
LaVAN \cite{karmon2018lavan} & individual, universal & visible & $L_2$ & white-box & targeted & gradient \\ Universal Adversarial Networks (UAN) \cite{hayes2018learning} & universal & optimal & $L_2$, $L_{\infty}$ & white-box & targeted & gradient \\ Expectation Over Transformation (EOT) \cite{Athalye2017} & individual & optimal & $L_2$ & white-box & targeted & gradient \\ Local Search Attack (LSA) \cite{narodytska2017simple} & individual & optimal & $L_0$ & black-box & targeted, untargeted & gradient \\ Natural Evolutionary Strategies (NES) \cite{ilyas2018black} & individual & optimal & $L_\infty$ & black-box & targeted & approximation \\
Boundary Attack (BA) \cite{brendel2017decision} & individual & optimal & $L_2$ & black-box & targeted, untargeted & decision \\
CW Attack \cite{carlini2017towards} & individual & optimal & $L_0$, $L_2$, $L_\infty$ & white-box & targeted, untargeted & gradient \\
GenAttack \cite{alzantot2018genattack} & individual & optimal & $L_2$, $L_\infty$ & black-box & targeted & decision \\
BIM and ILCM \cite{Kurakin2016a} & individual & optimal & $L_\infty$ & white-box & untargeted & gradient \\
Momentum Iterative Method (M-BIM) \cite{dong2018boosting} & individual & optimal & $L_\infty$ & white-box, black-box & untargeted & gradient \\
Zeroth-Order Optimization (ZOO) \cite{chen2017zoo} & individual & optimal & $L_2$ & black-box & targeted, untargeted & transfer, score \\ Hot-Cold Attack \cite{rozsa2016adversarial} & individual & optimal & $L_2$ & white-box & targeted & gradient \\
Projected Gradient Descent (PGD) \cite{madry2017towards} & individual & optimal & $L_1$, $L_\infty$ & white-box & targeted & gradient \\
UPSET \cite{sarkar2017upset} & universal & optimal & $L_2$ & black-box & targeted & gradient \\
ANGRI \cite{sarkar2017upset} & individual & optimal & $L_2$ & black-box & targeted & gradient \\
Elastic-Net Attack (EAD) \cite{chen2018ead} & individual & optimal & $L_1$ & white-box & targeted, untargeted & gradient \\
Hop-Skip-Jump Attack (HSJ) \cite{hopskipjump} & individual & optimal & $L_2$, $L_\infty$ & black-box & targeted, untargeted & decision \\
Robust Physical Perturbations (RP2) \cite{eykholt2017robust} & individual & physical & $L_1$, $L_2$ & white-box & targeted & gradient \\
Ground-Truth Attack \cite{ground_truth} & individual & optimal & $L_1$, $L_\infty$ & white-box & targeted & gradient \\ OptMargin \cite{he2018decision_opt} & individual & optimal & $L_01$, $L_2$, $L_\infty$ & white-box & targeted & gradient \\
One-Pixel Attack \cite{su2019onepixel} & individual & visible & $L_0$ & black-box & targeted, untargeted & decision \\ BPDA \cite{athalye2018obfuscated} & individual & optimal & $L_2, L_\infty$ & black-box & untargeted, targeted & approximation \\ SPSA \cite{uesato2018adversarial} & individual & optimal & $L_\infty$ & black-box & untargeted & approximation \\
Spatially Transformed Network (stAdv) \cite{xiao2018spatially} & individual & optimal & custom & white-box & targeted & gradient \\ AdvGAN \cite{xiao2018generating} & individual & optimal & $L_2$ & grey-box, black-box & targeted & gradient\\
Houdini \cite{cisse2017houdini} & individual & optimal & $L_2$, $L_\infty$ & black-box & targeted & gradient \\
Adversarial Transformation Networks (ATNs) \cite{baluja2017adversarial} & individual & optimal & $L_\infty$ & white-box & targeted & gradient \\ \bottomrule
\end{tabular}
\label{tbl:attacks}
\begin{tablenotes}
  \normalsize
  \item[*]The axis \textit{Attacker's Influence} is not present in Table \ref{tbl:attacks} because it does not depend of any aforementioned attack algorithm. Similarly, the axis \textit{Attack Computation} is not also present because, except from FGSM and L-BFGS, all other attacks mentioned in Table \ref{tbl:attacks} have their respective perturbations computed iterativelly.
\end{tablenotes}}
\end{threeparttable}
\end{adjustbox}
\vspace{-5mm}
\end{table}

%% file: sections/sec4.tex
The menace of adversarial images has encouraged the scientific community to elaborate several approaches to defend classification models. However, design such countermeasures has shown to be a difficult task once adversarial inputs are solutions to an optimization problem that is non-linear and non-convex. Since good theoretical tools for describing the solutions to these optimization problems do not exist, it is very hard to put forward a theoretical argument ensuring a defense strategy will be efficient against adversarial examples \cite{Kumar2017}. Therefore, the existing defense mechanisms have some limitations in the sense that they can provide robustness against attacks in specific threat models. The design of a robust machine learning model against all types of adversarial images and other examples is still an open research problem \cite[p.~27]{chakraborty2018adversarial}. 

\subsection{Taxonomy of Defenses Against Adversarial Attacks} \label{subsec:def_tax}

This section categorizes the defenses against adversarial attacks using a novel taxonomy composed of two different axes, namely \textit{(i) defense objective} and \textit{(ii) defense approach}.

\subsubsection{Defense Objective}
According to its main objetive, a defense can be \textit{(i) proactive} or \textit{(ii) reactive}. Proactive defenses aim to turn classification models more robust to adversarial images. A model is considered robust when it is able to correctly classify an adversarial image as if it were a legitimate image. On the other hand, reactive defenses focus on detecting adversarial images by acting as a filter that identifies malicious images before they reach the classifier. The detected images are usually either discarted or sent to a recovery procedure.

\subsubsection{Defense Aproach} \label{subsub:adv_training}

Defenses can adopt different approaches when protecting models against adversarial images. Each approach groups a set of similar procedures, which can range from brute force solutions to preprocessing techniques. Based on a systematic review of literature, this paper also categorizes the most relevant proactive and reactive countermeasures according to their operational approach, which can be: \textit{(i) gradient masking}, \textit{(ii) auxiliary detection models}, \textit{(iii) statistical methods}, \textit{(iv) preprocessing techniques}. \textit{(v) ensemble of classifiers} and \textit{(vi) proximity measurements}. 

\paragraph{\textbf{Gradient Masking:}} defenses based on gradient masking (effect also known as \textit{obfuscated gradient} \cite{athalye2018obfuscated})  produce, sometimes unintentionally, models containing smoother gradients that hinders optimization-based attack algorithms from finding wrong directions in space, \textit{i.e.} without useful gradients for generating adversarial examples. According to \citeauthor{athalye2018obfuscated} \cite{athalye2018obfuscated}, defenses based on gradient masking can be organized in: \textit{(i) shattered gradients, (ii) stochastic gradients} and \textit{(iii) exploding/vanishing gradients}.

\begin{itemize}[leftmargin=*]
    \item \textit{Shattered gradients:} are caused by non-differentiable defenses, thus introducing nonexistent or incorrect gradients;
    
    \item \textit{Stochastic gradients:} are caused by randomized proactive/reactive defenses or randomized preprocessing on inputs before being fed to the classifier. This strategy of gradient masking usually leads an adversarial attack to incorrectly estimate the true gradient;
    
    \item \textit{Exploding/vanishing gradients:} are caused by defenses formed by very deep architectures, usually consisting of multiple iterations of a neural network evaluation, where the output of one layer is fed as input of the next layer.
\end{itemize}

\begin{figure*}[h!]
	\centering
	\includegraphics[width=0.7\textwidth]{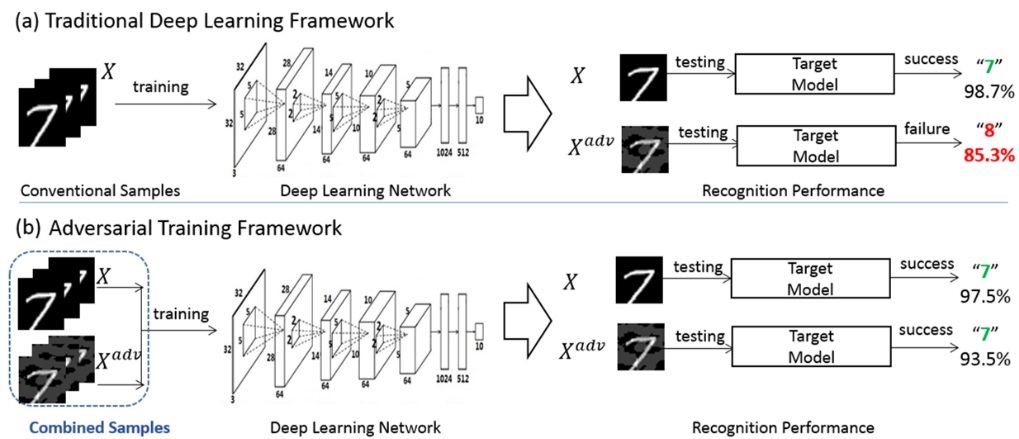}
	\caption{Adversarial training increases the robustness of classifiers by training them using an augmented training dataset containing adversarial images. Adapted from \citeauthor{shen2017ape} \cite{shen2017ape}.}
	\label{fig:adv_train}
\end{figure*}

Basically, there are many countermeasures based on different strategies of gradient masking, as can be seen in Table \ref{tbl:countermeasures}. However, two distinct strategies are frequently mentioned by related work in literature, which in turn make them relevant to describe in more details: \textit{(i) Adversarial Training} and \textit{(ii) Defensive Distillation}. 

Defenses based on adversarial training are usually considered in literature a brute force approach to protect against adversarial examples. Essentially, the main objetive of adversarial training is to make a classification model more robust by training it in a dataset containing legitimate and adversarial images. Formally, given a tuple $X = (x, y)$, where $x$ is a legitimate image, $y$ the class $x$ belongs to and $T$ a training dataset having only the tuple $X$\footnote{For didactic purposes, consider the training dataset $T$ formed by only one image.}, such as $T = \{X\}$, an adversarial image $x'$ is crafted from $x$ by an attack algorithm $A$, thus forming a new tuple $X'$ that will have the same label $y$ of the clean image $x$, such that $X' = \{x', y\}, x' = A(x)$. Afterwards, the training dataset $T$ is augmented with $X'$ and now contains two image tuples: $T' = \{X, X'\}$. The learning model is then retrained using the training dataset $T'$, resulting in a theoretically stronger model (see Figure \ref{fig:adv_train}).

Despite the good results adversarial training has presented in several works \cite{Szegedy2013, Goodfellow2014, Huang2015, Kurakin2016, Tramer2017a, Zantedeschi2017, madry2017towards, kannan2018adversarial}, this gradient masking approach has basically two issues. The first issue is related to the strong coupling adversarial training has with the attack algorithm used during the training process. Retraining a model with adversarial training does not produce a generic model which is able to resist against evasions of adversarial images generated by a different attack algorithm not used in the training process. In order to have a more generic model, it would be necessary to elaborate a training dataset $T$ with a massive amount of adversarial images generated using different attack algorithms and amounts of disturbance. Therefore, the second issue concerning adversarial training raises: this is a procedure computationally inefficient, given two facts: (i) the great number of adversarial images that must be crafted from different attacks, which in turn does not guarantee robustness against adversarial images generated from more complex algorithms and (ii) after generating these malicious images, the model must train using a much larger dataset, which exponentially grows the training time. A robust defense method must be decoupled from any attack algorithm to increase its generalization. Notwithstanding the drawbacks, \citeauthor{madry2017towards} \cite{madry2017towards} proposed training on adversarial samples crafted using Projected Gradient Descent (PGD) attack which is, by the time of this writing, the most promising defense present in literature, since it has shown robustness to various types of attacks in both white-box and black-box settings \cite{wiyatno2019adversarial}. However, their method is not model-agnostic and,
due to computational complexities, it has not been tested
on large-scale datasets such as ImageNet \cite{moosavi2018D3}.

\begin{figure*}[h!]
	\centering
	\includegraphics[width=0.75\textwidth]{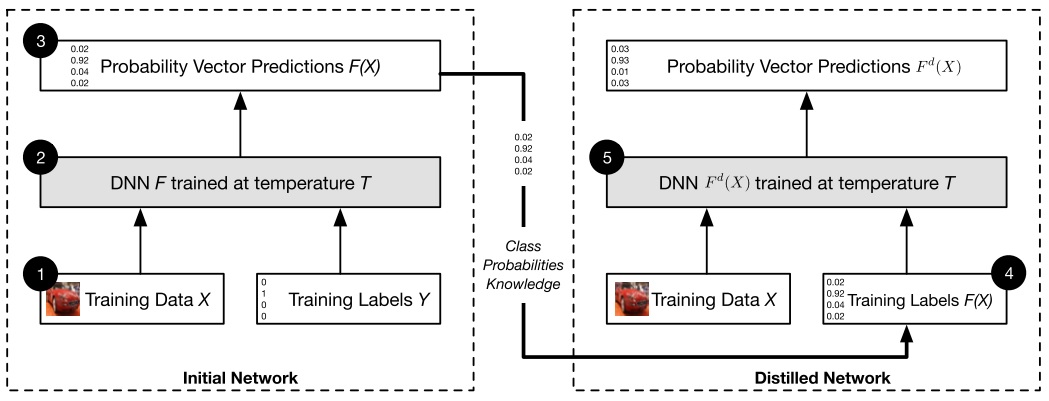}
	\caption{Schematic model of defensive distillation \cite{Papernot2016b}.}
	\label{fig:defensive_dist}
	\vspace{-3mm}
\end{figure*}

Defensive Distillation, in turn, is a proactive defense initially proposed by \citeauthor{Papernot2016b} \cite{Papernot2016b}. This countermeasure is inspired by a technique based on transfer of knowledge among learning models known as distillation \citep{hinton2015distilling}. In learning distillation, the knowledge acquired by a complex model, after being trained using a determined dataset, is transfered to a simpler model. In a similar way, defensive distillation firstly trains a model $f$ using a dataset containing samples $X$ and labels $Y$ with a temperature $t$, resulting as output a probabilistic vector $f(X)$. The label set $Y$ is then replaced by the probabilistic vector $f(X)$ and a model $f^d$ with the same architecture of $f$ is created and trained with the sample set $X$, but now using as labels the novel label set $f(X)$. By the end of training, the destilled probabilistic output $f^d(X)$ is produced. Figure \ref{fig:defensive_dist} depicts the schematic model of defensive distillation.

Defenses based on gradient masking usually produce models containing smoother gradients in certain regions of space, making harder for the attacker to find promising directions to perturb an image. However, the attacker can instead use an non-differentiable attack, such as BPDA \cite{athalye2018obfuscated} or SPSA \cite{uesato2018adversarial} as well as perform a black-box attack by training a surrogate model. This surrogate model reproduces the behaviour of the targeted model, since the attacker queries it using images carefully crafted by him and watches the outputs the targeted model gives. Then, the attacker takes advantage of the transferibility property of adversarial examples by using the gradients of the surrogate model in order to craft the images that will also lead the target model to misclassifications \cite{gradient_masking}. Section \ref{subsec:transferability} gives more information regarding the transferability property of adversarial examples.

\paragraph{\textbf{Auxiliary Detection Models (ADMs)}}

A defense based on ADMs is, usually, a reactive method that, basically, makes use of adversarial training to elaborate an auxiliary binary model that will act as a filter after being trained, checking whether an input image is legitimate or adversarial before sending it to the application classifier $f$. Works such as \citeauthor{Grosse2017, Gong2017, Metzen2017} and \citeauthor{chen2017reabsnet} \cite{Grosse2017, Gong2017, Metzen2017, chen2017reabsnet} have proposed defenses based on ADMs.

\citeauthor{Grosse2017} \cite{Grosse2017} have adapted an application classifier $f$ to also act as a ADM, training it in a dataset containing $n+1$ classes. The procedure followed by the authors consists of generating adversarial images $x'_i$ for each legitimate image $(x_i, y_j)$ that belongs to the training set $\mathcal{T}$, where $i \leq |{\mathcal{T}}| \times m$ (where $m$ is the number of attack algorithms used) and $j \leq n$. After the generation of adversarial images, it has been formed a new training set $\mathcal{T}_1$, where $\mathcal{T}_1 = \mathcal{T} \cup \{(x'_i, n + 1), i \leq |{\mathcal{T}}| \times m\}$. $n + 1$ is the label assigned to an adversarial image. Finally, the model $f$ has been trained using the $\mathcal{T}_1$ set.

\citeauthor{Gong2017} \cite{Gong2017} have elaborated a defense similar to \citeauthor{Grosse2017}, but instead of adapting the application classifier to predict adversarial images in a class $n+1$, the authors have built and trained an ADM to filter out adversarial images $X'$ (crafted by FGSM and JSMA attacks) from the legitimate images $X$, using a training dataset $\mathcal{T}_1$, formed from $\mathcal{T}$. Formally, $\mathcal{T}_1 = \{(x_i, 1) : i \in {|\mathcal{T}|}\} \cup \{(x'_i, 0) : i \leq {|\mathcal{T}|} \times m\}$.

In \citeauthor{Metzen2017} \cite{Metzen2017}, the representation outputs of the hidden layers of a DNN have been used for training some ADMs, in a way similar of what has been made in \cite{Gong2017}. The authors have named these ADMs \textit{subnetworks} and fixed them among specific hidden layers of a DNN in order to detect adversarial images. In this work were performed experiments using the attack algorithms FGSM, BIM, and DeepFool.

Finally, \citeauthor{chen2017reabsnet} \cite{chen2017reabsnet} have elaborated a detection and reforming architecture called \textit{ReabsNet}. When Reabsnet receives an image $x$, it uses an ADM (represented by a DNN trained by adversarial training) to check whether $x$ is legitimate or adversarial. In case of being classified as legitimate by the ADM, Reabsnet sends $x$ to the application classifier. However, in case of being classified as adversarial, $x$ is sent by ReabsNet to an iterative process, that reforms the image $x$ while it is classified by adversarial by the ADM. In the end of the reform process, the image $x$ is finally sent to the application classifier.

\paragraph{\textbf{Statistical Methods:}}

Some works such as \citeauthor{Grosse2017} and \citeauthor{Feinman2017} \cite{Grosse2017, Feinman2017} have performed statistical comparisons among the distributions of legitimate and adversarial images. \citeauthor{Grosse2017} have elaborated a reactive defense method that has performed an approximation for the hypothesis test MMD (\textit{Maximum Mean Discrepancy}) with the Fisher's permutation test in order to verify whether a legitimate dataset $\mathcal{S}_1$ belongs to the same distribution of another dataset $\mathcal{S}_2$, which may contain adversarial images. Formally, given two datasets  $\mathcal{S}_1$ and $\mathcal{S}_2$, it is initially defined $a = \text{MMD}(\mathcal{S}_1, \mathcal{S}_2)$. Later, there is a permutation of elements of $\mathcal{S}_1$ and $\mathcal{S}_2$ in two new datasets $\mathcal{S}'_1$ and $\mathcal{S}'_2$, and it is defined $b = \text{MMD}(\mathcal{S}'_1, \mathcal{S}'_2)$. If $a < b$, the null hypothesis is rejected and then it is concluded that the two datasets belong to different distributions. This process is repeated several times and the \textit{p-value} is defined as the fraction of the number of times which the null hypothesis was rejected.

\citeauthor{Feinman2017} \cite{Feinman2017} have also proposed a reactive defense called \textit{Kernel Density Estimation} (KDE). KDE makes use of Gaussian Mixture Models\footnote{Gaussian Mixture Models are unsupervised learning models that clusters data by representing sub-populations, using normal distributions, within a general population.} to analyze the outputs of the logits layer of a DNN and to verify whether the input images belong to the same distribution of legitimate images. Given an image $x$ classified as a label $y$, the KDE estimates the probability of $x$ according to Equation \ref{eq:kde}:

\begin{equation}\label{eq:kde}
    \text{KDE}(x) = \frac{1}{\abs{X_y}} \sum_{s \in X_y}{\exp\left({\frac{\abs{F^{n-1}(x) - F^{n-1}(s)}^2}{\sigma^2}}\right)}
\end{equation}

\noindent where $X_y$ is the training dataset containing images pertaining the class $y$ and $F^{n-1}(x)$ is the logits output $Z$ related to input $x$. Therefore, the detector is built by the selection of a threshold $\tau$ which classifies $x$ as adversarial if $KDE(x) < \tau$ or legitimate, otherwise.

\paragraph{\textbf{Preprocessing Techniques}}

Other works have elaborated countermeasures based on preprocessing techniques, such as image transformations \cite{Xie2017, guo2017countering}, GANs \cite{shen2017ape, samangouei2018defense}, noise layers \cite{liu2017towards}, {denoising autoencoders} \cite{Gu2014} and dimensionality reduction \cite{Hendrycks2017,Li2016adv,xu2017feature}. In the following, each work will be explained in more details.

\citeauthor{Xie2017} \cite{Xie2017} have elaborated a proactive defense called Random Resizing and Padding (RRP) that inserts a resizing and a padding layer in the beginning of a DNN architecture. The resizing layer alters the dimensions of a input image, and later, the padding layer inserts null values in random positions on the surrondings of the resized image. In the end of the padding procedure, the resized image is classified by the proactive model. 

\citeauthor{guo2017countering} \cite{guo2017countering} have applied various transformations in input images before classification, such as cropping and rescaling, bit-depth reduction, JPEG compression, total variance minimization (TVM) and image quilting. \citeauthor{guo2017countering} have implemented TVM as a defense by randomly picking pixels from an input and performing iterative optimization to find an image whose colors are consistent with the randomly picked pixels. On the other hand, image quilting has involved reconstructing an image using small patches taken from the training database by using a nearest neighbor procedure (e.g. kNN). The intuition behind image quilting is to construct an image that is free
from adversarial perturbations, since quilting only uses clean patches to reconstruct the image \cite{wiyatno2019adversarial}. The authors claimed that TVM and image quilting have presented the best results when protecting the classifier since both (i) introduce randomness, (ii) are non-differentiable operations which hinders the attacker to compute the model gradient and (iii) are model-agnostic which means the model does not need to be retrained or fine-tuned.

\citeauthor{shen2017ape} \cite{shen2017ape} have proposed a proactive defense method which have adapted a GAN to preprocess input images before they be sent to the application classifier. \citeauthor{samangouei2018defense} \cite{samangouei2018defense} have also elaborated a defense based on a GAN framework that uses
a generative transformation network $G$ which projects an input image $x$ onto the range of the generator by minimizing the reconstruction error $||G(z)-x||^2$. After the transformation, the classifier is fed with the reconstruction $G(z)$. Since the generator was trained to model the unperturbed training data distribution, the authors claimed this added step results in a substantial reduction of any potential adversarial noise. In turn, \citeauthor{liu2017towards} \cite{liu2017towards} have adopted an approach based on noise layers. {These noise layers have been inserted among the hidden layers of a CNN in order to apply a gaussian noise randomly crafted on each vector of the input image. According to the authors, this procedure avoids gradient-based attacks.} \citeauthor{Gu2014} \cite{Gu2014} have elaborated the Deep Contractive Networks (DCNs), which are proactive defense methods that make use denoising autoencoders and evolutionary algorithms as alternatives to remove perturbations from adversarial images.

In addition, there are countermeasures that preprocesses input images using dimensionality reduction techniques \cite{Hendrycks2017, Li2016adv, xu2017feature}. These works are based on the hypothesis that, by reducing the dimensions of an input, the likelihood of an attacker creating a perturbation that can affect the classifier's performance decreases, given the fact the attack algorithm will have less information concerning the hyperspace of the image \cite{xu2017feature}. Keeping this hypothesis in mind, \citeauthor{Hendrycks2017} \cite{Hendrycks2017} have elaborated a reactive defense based on Principal Component Analysis (PCA)\footnote{PCA is a dimensionality reduction technique that reduces, by appling a linear transformation, a set of points in $n$-dimensional space to a $k$-dimensional space, where $k \leq n$.}. The authors have inferred that adversarial images assign greater weights on larger principal components and smaller weights on initial principal components. \citeauthor{Li2016adv} \cite{Li2016adv} have applied PCA on values produced by the convolution layers of a DNN and then used a cascate classifier to detect adversarial images. The cascate classifier $C$ classifies an image $x$ as legitimate only if all its subclassifiers $C_i$ classify $x$ as legitimate, but rejects $x$ if some classifier $C_i$ reject $x$. In this work, the L-BFGS attack has been used to perform the experiments. 

\begin{figure}[h!]
	\centering
	\includegraphics[width=0.65\textwidth]{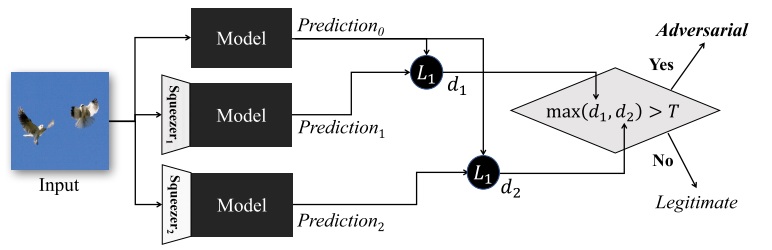}
	\caption{The Feature Squeezing workflow \cite{xu2017feature}.}
	\label{fig:squeezing}
	\vspace{-3mm}
\end{figure}

\citeauthor{xu2017feature} \cite{xu2017feature} have introduced \textit{Feature Squeezing}, which is reactive defense that makes use of two techniques to reduce de dimensionality of a input image: (i) color bit reduction and (ii) spatial smoothing. According to the authors, these techniques have been chosen since they complement each other by treating two different types of perturbation. The bit reduction aims to eliminate small perturbations by covering various pixels, while spatial smoothing aims to eliminate big perturbations by covering some pixels. During the detection process, Feature Squeezing generates two reduced versions of an input image $x$: (i) $\hat{x}_1$, which represents image $x$ with the color bits reduced and (ii) $\hat{x}_2$, which represents $x$ reduced with spatial smoothing. Later, \textit{Feature Squeezing} sends the images $x$, $\hat{x}_1$ e $\hat{x}_2$ to be classified by a DNN $f$ and compares the softmax outputs $f(x)$, $f(\hat{x}_1)$ e $f(\hat{x}_2)$ using the $L_1$ metric. If the $L_1$ metric exceeds a predefined threshold $\tau$, Feature Squeezing classifies $x$ as an adversarial example and discarts it. Figure \ref{fig:squeezing} depicts this workflow. 

\paragraph{\textbf{Ensemble of Classifiers}}

Defenses based on ensemble of classifiers are countermeasures formed by two or more classification models that can be chosen in runtime. This approach is based on the assumption that each model reciprocally compensates the weaknesses other model eventually might have when classifying a given input image \cite{He2017}. Works such as \citeauthor{strauss2017ensemble, Tramer2017a, abbasi2017robustness} and \citeauthor{mtdeep2017} \cite{strauss2017ensemble, Tramer2017a, abbasi2017robustness, mtdeep2017} have adopted different techniques to elaborate defenses based on ensemble of classifiers. \citeauthor{mtdeep2017} \cite{mtdeep2017} have used a bayesian algorithm to chose a optimal model from an ensemble so as to minimize the chances of evasion and, at the same time, maximize the correct predictions on legitimate images. \citeauthor{abbasi2017robustness} \cite{abbasi2017robustness} have formed ensembles of specialist models which detects and classifies an input image by majority vote. \citeauthor{strauss2017ensemble} \cite{strauss2017ensemble} have made empirical evaluations based on four types of different ensembles and trainings. \citeauthor{Tramer2017a} \cite{Tramer2017a}, in turn, have used a variation of adversarial training to train the main classifier with adversarial images crafted by an ensemble of DNNs.

\paragraph{\textbf{Proximity Measurements}}

\begin{figure}[h!]
	\centering
	\includegraphics[width=0.6\textwidth]{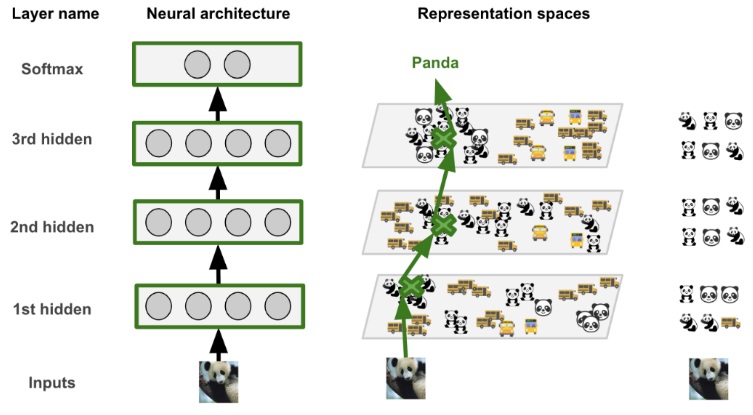}
	\caption{The DkNN computes uncertainty and reliability metrics to support a prediction made by a DNN for a given input image $x$, by performing a search among the training images with internal representations closest to $x$ \cite{papernot2018marauder}.}
	\label{fig:dknn}
\end{figure}

There are other works such as \citeauthor{papernot2018deepnearest, Cao, carrara2018adversarial, Meng2017, machado_iceis2019} which have proposed defenses based on proximity measurements among legitimate and adversarial images to the decision boundary. \citeauthor{papernot2018deepnearest} \cite{papernot2018deepnearest} have elaborated an proactive defense method called \textit{Deep k-Nearest Neighbors} (DkNN), which makes use of a variation of the kNN algorithm\footnote{kNN stands for k-Nearest Neighbors. It is an supervised classification algorithm that assigns, to a given input $x$, the most frequent class $c$ among the $k$ nearest training samples from $x$, according to a certain distance metric.} to compute uncertainty and reliability metrics from the proximity among the hidden representations of training and input images, obtained from each layer of a DNN. The labels representing points in space of training images are analyzed after the input image goes through all the layers of the DNN. In case of the prediction related to the input $x$, given by the DNN, be in accordance with the labels representing the nearest training images to $x$, the uncertainty metric tends to be small. In contrast, in case of the labels of the training images be divergent among them, the uncertainty metric tends to be large \cite{papernot2018marauder}. Figure \ref{fig:dknn} depicts the DkNN operation.

\citeauthor{Cao} \cite{Cao} have also adopted an approach based on proximity metrics to elaborate a proactive countermeasure, called \textit{Region-based Classification} (RC). RC is a variation of the kNN which defines a region $R$ in hyperspace, having as centroid an input image $x$, assigning it a label corresponding to the class which most intersects the area of this region $R$. Formally, for a given input image $x$ and a DNN $f$ that splits the hyperspace in $C$ distinct regions $R = \{R_1, R_2, \cdots, R_C\}$ (being $C$ the number of classes and $R_i$ the predicted class for $f(x)$), it is created a hypercube $B(x,r)$ around $x$ (being $x$ the centroid of $B(x,r)$) with length $r$. $A_i(B(x, r))$ is the area of hypercube $B(x,r)$ that intersects the region $R_i$. The classifier $RC$ formed from $f$ is defined as $RC_{f,r}$ and its prediction for $x$ is based on the region $R_i$ which has the largest intersection with the area of hypercube, namely $RC_{f, r} = \text{argmax}_i({A_i(B(x, r))})$.

\citeauthor{carrara2018adversarial} \cite{carrara2018adversarial} have introduced a reactive defense that resembles somehow the DkNN architecture \cite{papernot2018deepnearest}. The reactive method proposed by \citeauthor{carrara2018adversarial}, at first, make use of a DNN $f$ to classify an input image $x$. {Afterwards}, the inner representations regarding image $x$, obtained from a hidden layer of $f$ (such layer is chosen empirically), are used by an kNN algorithm to perform a search in the training dataset in order to recover the $k$ images containing the most similar representations to the corresponding representations of $x$. Therefore, it is obtained a confidence metric $conf$ related to the prediction $f(x)$, where $conf$ is computed based on the score of the kNN algorithm. If the confidence is below a predefined threshold, the input image is classified as adversarial and discarted afterwards; otherwise, the prediction $f(x)$ is valid with a confidence level of $c$.

In turn, \citeauthor{Meng2017} \cite{Meng2017} have proposed MagNet: a non-deterministic and reactive architecture composed of two defense layers: (i) a detection layer, which rejects adversarial images containing large perturbations and, for this reason, considered further from the decision boundary, and (ii) the reform layer, which reforms the images derived from the detection layer as an attempt to remove any existing perturbations that are still present in them. According to the authors, the reform layer acts as a "magnet", drawing the adversarial images that evaded the detection layer to the regions of the decision boundary corresponding to their respective correct classes. For both layers, MagNet randomly chooses from a repository two defense components, implemented as autoencoders (trained beforehand using legitimate images): one autoencoder for the detection layer and the other for the reformer layer. The non-deterministic choice of the components is, according to the authors, inspired on cryptography techniques to reduce the chances of evasions.

In \citeauthor{vorobeychik2018adversarial} \cite{vorobeychik2018adversarial} is said that a defense based on randomness may be an inportant strategy to secure machine learning algorithms. Since randomness can significantly increase the size of perturbations and the computational cost needed to craft adversarial images, \citeauthor{machado_iceis2019} \cite{machado_iceis2019} have extended the non-deterministic effect of MagNet by proposing a defense called MultiMagNet, which randomly chooses multiple defense components at runtime instead of just one, how is originally made by MagNet. In a way similar to MagNet, the MultiMagNet's defense components have also been implemented as autoencoders trained on
legitimate images. Later, the authors have split the MultiMagNet's architecture into two stages, namely \textit{(i) calibration stage} and \textit{(ii) deployment stage}. In calibration stage, MultiMagNet makes use of a validation dataset to find the best set of hyperparameters. Once calibrated, MultiMagNet goes to the \textit{(ii) deployment stage}, where it analyzes input images in order to protect the application classifier against adversarial examples. The authors have made a comparative study with MagNet using legitimate and adversarial images crafted by FGSM, BIM DeepFool and CW attacks, and concluded the increasing of the non-deterministic effect by choosing multiple components can lead to better defense architectures.

Table \ref{tbl:countermeasures} makes a comprehensive overview of some relevant defenses against adversarial attacks in Computer Vision available in literature, following the taxonomy presented in Section \ref{subsec:def_tax}. It also shows which of them have already been circumvented by mentioning the corresponding works on adversarial attacks.

\begin{table}[h!]
\centering
\caption{Summary of some relevant defenses against adversarial attacks in Computer Vision.}
\begin{adjustbox}{max width=\textwidth}
\begin{threeparttable}
\begin{tabular}{@{}lccccc@{}}
\toprule
\multirow{2}{*}{\begin{tabular}[c]{@{}c@{}}Defense / Work\\and Reference\end{tabular}} & \multirow{2}{*}{Objective} & \multirow{2}{*}{Approach} & \multicolumn{2}{c}{Robustness Claims} & \multirow{2}{*}{Bypassed by \tnote{**}} \\ \cmidrule(lr){4-5}
 &  &  & Attack algorithms & Attacker's knowledge\tnote{*} &  \\ \midrule
Thermometer Encoding \cite{buckman2018thermometer} & Proactive & Preprocessing & PGD & WB, BB & \citeauthor{athalye2018obfuscated} \cite{athalye2018obfuscated} \\
VectorDefense \cite{kabilan2018vectordefense} & Proactive & Preprocessing & BIM, JSMA, DeepFool, CW, PGD & WB, GB & \textemdash \\
PixelDefend \cite{song2017pixeldefend} & Proactive, Reactive & Preprocessing, Proximity & FGSM, BIM, DeepFool, CW & WB & \citeauthor{athalye2018obfuscated} \cite{athalye2018obfuscated} \\ \citeauthor{mustafa2019SuperResolution} \cite{mustafa2019SuperResolution} & Proactive & Preprocessing & \begin{tabular}[c]{@{}c@{}}FGSM, BIM, MI-BIM, DeepFool, CW\end{tabular} & WB, BB & \textemdash \\
\citeauthor{prakash2018deflecting} \cite{prakash2018deflecting} & Proactive & Preprocessing & \begin{tabular}[c]{@{}c@{}}FGSM, BIM, JSMA, DeepFool, L-BFGS, CW\end{tabular} & WB & \citeauthor{athalye2018robustness} \cite{athalye2018robustness} \\
SAP \cite{dhillon2018stochastic} & Proactive & Gradient Masking & FGSM & WB & \citeauthor{athalye2018obfuscated} \cite{athalye2018obfuscated} \\
\citeauthor{Feinman2017} \cite{Feinman2017} & Reactive & Statistics & FGSM, BIM, JSMA, CW & WB & \citeauthor{carlini2017bypassing} \cite{carlini2017bypassing} \\
\citeauthor{carrara2018adversarial} \cite{carrara2018adversarial} & Reactive & Proximity & L-BFGS, FGSM & WB & \textemdash \\ D3 algorithm \cite{moosavi2018D3} & Proactive & Preprocessing & FGSM, DeepFool, CW, UAP & \begin{tabular}[c]{@{}c@{}}WB BB, GB\\\end{tabular} & \textemdash \\
RRP \cite{Xie2017} & Proactive & Preprocessing & FGSM, DeepFool, CW & WB & \citeauthor{uesato2018adversarial} \cite{uesato2018adversarial} \\
RSE \cite{liu2017towards} & Proactive & Preprocessing, Ensemble & CW & WB, BB & \textemdash \\ \citeauthor{Bhagoji2017} \cite{Bhagoji2017} & Proactive & Preprocessing & FGSM & WB & \citeauthor{carlini2017bypassing} \cite{carlini2017bypassing}  \\
\citeauthor{li2016adversarial} \cite{li2016adversarial} & Reactive & Preprocessing, Statistics & L-BFGS & WB & \citeauthor{carlini2017bypassing} \cite{carlini2017bypassing} \\
ReabsNet \cite{chen2017reabsnet} & Reactive & ADM, Preprocessing & FGSM, DeepFool, CW & WB & \textemdash \\ \citeauthor{zheng2018robust} \cite{zheng2018robust} & Reactive & Statistics, Proximity & FGSM, BIM, DeepFool & BB, GB & \textemdash \\ DeT \cite{li2019det} & Proactive & Preprocessing, Ensemble & FGSM, BIM, DeepFool, CW & BB, GB & \textemdash \\ Deep Defense \cite{yan2018deep} & Proactive & Gradient Masking & DeepFool & WB & \textemdash \\
\citeauthor{Grosse2017} \cite{Grosse2017} & Reactive & Statistics & FGSM, JSMA & WB, BB & \citeauthor{carlini2017bypassing} \cite{carlini2017bypassing} \\
RCE \cite{pang2017RCE} & Reactive & Gradient Masking & FGSM, BIM, ILCM, JSMA, CW & WB, BB & \textemdash \\
NIC \cite{ma2019nic} & Reactive & ADM, Proximity & FGSM, BIM, JSMA,  DeepFool, CW & WB, BB & \textemdash \\
\citeauthor{Cao} \cite{Cao} & Proactive & Proximity & FGSM, BIM, JSMA, DeepFool, CW & WB & \citeauthor{he2018decision_opt} \cite{he2018decision_opt} \\
\citeauthor{Hendrycks2017} \cite{Hendrycks2017} & Reactive & Preprocessing & FGSM & WB & \citeauthor{carlini2017bypassing} \cite{carlini2017bypassing} \\ Feature Distillation \cite{liu2018featureDistillation} & Proactive & Preprocessing & FGSM, BIM, DeepFool, CW, BPDA & WB, BB, GB & \textemdash \\
LID \cite{ma2018LID} & Reactive & Proximity & FGSM, BIM, JSMA, CW & WB & \citeauthor{athalye2018obfuscated} \cite{athalye2018obfuscated} \\
\citeauthor{cohen2019detecting} \cite{cohen2019detecting} & Reactive & Proximity & FGSM, JSMA, DeepFool, CW & WB & \textemdash \\
BAT \cite{wang2019bilateral} & Proactive & Gradient Masking & FGSM, PGD & WB & \textemdash \\ \citeauthor{madry2017towards} \cite{madry2017towards} & Proactive & Gradient Masking & PGD & WB, BB & \citeauthor{athalye2018obfuscated} \cite{athalye2018obfuscated}\tnote{***} \\
MALADE \cite{srinivasan2018MaLa_counterstrike} & Proactive & Preprocessing  & FGSM, PGD, M-BIM, EAD, BPDA, EOT, BA & WB, BB & \textemdash \\
S2SNet \cite{folz2018S2SNet} & Proactive & Gradient Masking & FGSM, BIM, CW & WB, GB & \textemdash \\
Gong et al. \cite{Gong2017} & Reactive & ADM & FGSM, JSMA & WB & \citeauthor{carlini2017bypassing} \cite{carlini2017bypassing} \\
Metzen et al. \cite{Metzen2017} & Reactive & ADM & FGSM, BIM, DeepFool & WB & \citeauthor{carlini2017bypassing} \cite{carlini2017bypassing} \\
\citeauthor{das2017jpegCompression} \cite{das2017jpegCompression} & Proactive & Preprocessing, Ensemble & FGSM, DeepFool & WB & \textemdash \\
CCNs \cite{ranjan2017CCN} & Proactive & Preprocessing & FGSM, DeepFool & WB, BB & \textemdash \\
DCNs \cite{Gu2014} & Proactive & Gradient Masking, Preprocessing & L-BFGS & WB & \textemdash \\ \citeauthor{na2017cascade} \cite{na2017cascade} & Proactive & Gradient Masking & FGSM, BIM, ILCM, CW & WB, BB & \textemdash \\
MagNet \cite{Meng2017} & Reactive & Proximity, Preprocessing & FGSM, BIM, DeepFool, CW & BB, GB & \citeauthor{carlini2017magnet} \cite{carlini2017magnet}  \\
MultiMagNet \cite{machado_iceis2019} & Reactive  & Proximity, Preprocessing, Ensemble & FGSM, BIM, DeepFool, CW & WB, BB. GB & \textemdash \\
WSNNS \cite{dubey2019WMM} & Proactive & Proximity & FGSM, CW, PGD & BB, GB & \textemdash \\
ME-Net \cite{yang2019menet} & Proactive & Preprocessing & FGSM, PGD, CW, BA, SPSA & WB, BB & \textemdash \\
SafetyNet \cite{lu2017safetynet} & Reactive & ADM & FGSM, BIM, JSMA, DeepFool & WB, BB & \textemdash \\
Defensive Distillation \cite{Papernot2016b} & Proactive & Gradient Masking & JSMA & WB & \citeauthor{carlini2017towards} \cite{carlini2017towards} \\
\citeauthor{Papernot2017ExtDefensive} \cite{Papernot2017ExtDefensive} & Proactive & Gradient Masking & FGSM, JSMA & WB, BB & \textemdash \\
Feature Squeezing \cite{xu2017feature} & Reactive & Preprocessing & FGSM, BIM, JSMA, CW & WB & \citeauthor{He2017} \cite{He2017}  \\
TwinNet \cite{ruan2018twinnet} & Reactive & ADM, Ensemble & UAP & WB & \textemdash \\
\citeauthor{abbasi2017robustness} \cite{abbasi2017robustness} & Reactive & Ensemble & FGSM, DeepFool & WB & \citeauthor{He2017} \cite{He2017}  \\
\citeauthor{strauss2017ensemble} \cite{strauss2017ensemble} & Proactive & Ensemble & FGSM, BIM & WB & \textemdash \\
\citeauthor{Tramer2017a} \cite{Tramer2017a} & Proactive & Gradient Masking, Ensemble & FGSM, ILCM, BIM & WB, BB & \citeauthor{alzantot2018genattack} \cite{alzantot2018genattack} \\
MTDeep \cite{mtdeep2017} & Proactive & Ensemble & FGSM, CW & WB & \textemdash \\
Defense-GAN \cite{samangouei2018defense} & Proactive & Preprocessing & FGSM, CW & WB, BB & \citeauthor{athalye2018obfuscated} \cite{athalye2018obfuscated} \\
APE-GAN \cite{shen2017ape} & Proactive & Preprocessing & L-BFGS, FGSM, DeepFool, JSMA, CW & WB &  \citeauthor{carlini2017magnet} \cite{carlini2017magnet} \\ \citeauthor{Zantedeschi2017} \cite{Zantedeschi2017} & Proactive & Gradient Masking & FGSM, JSMA, VAT \cite{miyato2016adversarial} & WB, BB & \citeauthor{carlini2017magnet} \cite{carlini2017magnet} \\
\citeauthor{NingHao2018ModelInterpretation} \cite{NingHao2018ModelInterpretation} & Reactive & Gradient Masking & FGSM, GDA \cite{biggio2013evasion}, POE \cite{wang2014man} & WB & \textemdash \\
\citeauthor{Liang2017} \cite{Liang2017} & Reactive & Preprocessing & FGSM, DeepFool, CW & WB & \textemdash \\
Parseval Networks \cite{cisse2017parseval} & Proactive & Gradient Masking & FGSM, BIM & BB & \textemdash \\ \citeauthor{guo2017countering} \cite{guo2017countering} & Proactive & Preprocessing & FGSM, BIM, DeepFool, CW & BB, GB & \citeauthor{dong2019evading} \cite{dong2019evading} \\
HGD \cite{liao2018HGD} & Proactive & Preprocessing & FGSM, BIM & WB, BB & \citeauthor{dong2019evading} \cite{dong2019evading} \\
ALP \cite{kannan2018adversarial} & Proactive & Gradient Masking & PGD & WB & \citeauthor{engstrom2018evaluating} \cite{engstrom2018evaluating} \\
\citeauthor{sinha2017WRM} \cite{sinha2017WRM} & Proactive & Gradient Masking & FGSM, BIM, PGD & WB & \textemdash \\
Fortified Networks \cite{lamb2019fortified} & Proactive & Preprocessing & FGSM, PGD & WB, BB & \textemdash \\ DeepCloak \cite{gao2017deepcloak} & Proactive & Preprocessing & L-BFGS, FGSM, JSMA & WB & \textemdash \\
\citeauthor{xie2019FeatureDenoisingBlock} \cite{xie2019FeatureDenoisingBlock} & Proactive & Preprocessing & FGSM, BIM, M-BIM, PGD & WB, BB & \citeauthor{kurakin2018adversarial} \cite{kurakin2018adversarial} \\ DDSA \cite{bakhti2019ddsa} & Proactive & Preprocessing & FGSM, M-BIM, CW, PGD & WB, BB, GB & \textemdash \\   ADV-BNN \cite{liu2018advbnn} & Proactive & Gradient Masking & PGD & WB, BB & \textemdash \\ DkNN \cite{papernot2018deepnearest} & Proactive & Proximity & FGSM, BIM, CW & WB & \citeauthor{sitawarin2019robustness} \cite{sitawarin2019robustness} \\ \hline
\end{tabular}
\begin{tablenotes}[flushleft]
  \Large
  \item[*]WB: White-box; BB: Black-box; GB: Grey-box.
  \item[**]The "\textemdash" symbol means that it has not been found in literature any work on adversarial attacks that has circunvented the respective defense.
  \item[***] Despite beign evaded in \citeauthor{athalye2018obfuscated} \cite{athalye2018obfuscated}, the method proposed by \citeauthor{madry2017towards} is considered as the state-of-the-art defense in literature \cite{athalye2018obfuscated}.
\end{tablenotes}
\end{threeparttable}
\end{adjustbox}
\label{tbl:countermeasures}
\vspace{-3mm}
\end{table}

%% file: sections/sec5.tex
Developing an understanding about the existence and the properties of adversarial examples, by reasoning why they affect the prediction of machine learning models, is usually the first step taken into consideration when elaborating attacks and defenses in Adversarial Machine Learning \cite{ma2018LID}. The vulnerability that CNNs and other machine learning algorithms present before the malicious effects of adversarial attacks is popularly known as \textit{Clever Hans Effect}, term somewhat popularized by the advent of CleverHans library \cite{papernot2018cleverhans}. This effect has been named after a german horse called Hans. His owner used to claim Hans has owned intellectual abilities by answering arithmetic questions that people made to it by tapping its hoof the number of times corresponding to the correct answer. However, after several experiments conducted on Hans, psychologists have concluded in fact the horse has not been solving arithmetic questions, but somehow it has developed the ability to identify behavioural signals made by the crowd, such as clappings and yielings, that warned it out to stop hitting its hoof on the ground. In other words, Hans has not developed an adaptive intelligence, but actually means of perceiving and interpreting its surroundings in order to correctly answer the questions.



Similar to Hans, learning models are usually able to give correct answers to complex problems, such as image recognition and classification, but without really learning from training data, what make them susceptible to adversarial attacks \cite{gershgorn2016fooling, Kumar2017}. Despite the absence of an unanimous accepted explanation for the adversarial paradox\footnote{\citeauthor{Tanay2016} \cite{Tanay2016} have defined this paradox as the disparity between high performance classification of state-of-the-art deep learning models against their susceptibility to small perturbations that differ so close from one class to another.}, this section will describe some common hypotheses present in literature regarding the existence of adversarial images. 

\subsection{High Non-Linearity Hypothesis}

\citeauthor{Szegedy2013} \cite{Szegedy2013} firstly concerned about the existence of adversarial examples. The authors have argued that adversarial examples exist due to the high non-linearity of deep neural networks, what contributes to the formation of low probability pockets in the data manifold that are hard to reach by sampling an input space around a given example (see Figure \ref{fig:tilting}a). According to \citeauthor{Gu2014} and \citeauthor{song2017pixeldefend} \cite{Gu2014, song2017pixeldefend}, the emergence of such pockets is given chiefly due to some deficiencies of objective functions, training procedures and datasets limited in size and diversity of training samples, thus leading models to poor generalizations.

\subsection{Linearity Hypothesis}\label{subsec:linear_hyp}

\citeauthor{Goodfellow2014} \cite{Goodfellow2014} contradicted the non-linearity hypothesis of \citeauthor{Szegedy2013} by assuming DNNs have a very linear behaviour caused by several activation functions like ReLU and sigmoid that perpetuates small perturbed inputs in a same wrong direction. As an attempt to underlie their explanation, the authors have elaborated the FGSM attack. \citeauthor{fawzi2015fundamental} \cite{fawzi2015fundamental} said that the robustness of a classifier is independent of the training procedure used and the distance between two classes is larger in high-order classifiers than in linear ones, suggesting that it is harder to find adversarial examples in deeper models. This explanation also goes against the non-linearity hypothesis of \citeauthor{Szegedy2013}. However, in contrast to the linearity hypothesis, \citeauthor{tabacof2016exploring} \cite{tabacof2016exploring} has found evidences the phenomenon of adversarial paradox may be a more complex problem, since results obtained from empirical experiments have suggested that shallow classifiers present a greater susceptibility to adversarial examples than deeper models. Despite some works that criticize the linearity hypothesis, some relevant attacks (such as FGSM \cite{Goodfellow2014} and DeepFool \cite{Moosavi-Dezfooli2015}) and defenses (such as Thermometer Encoding \cite{buckman2018thermometer}) have been based on it.

\subsection{Boundary Tilting Hypothesis}

\begin{figure}[h!]
    \centering
    \includegraphics[scale=0.375]{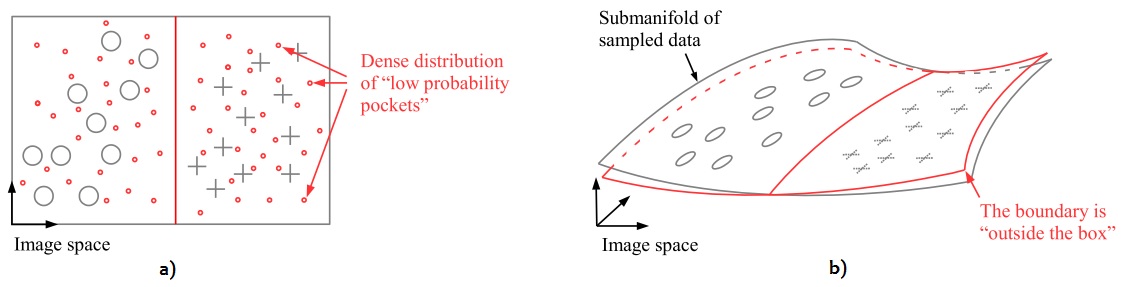}
    \caption{Comparison between the \citeauthor{Szegedy2013} and \citeauthor{Tanay2016}'s hypotheses \cite{Tanay2016}. a) \citeauthor{Szegedy2013}'s hypothesis lies on the assumption that an image space is densely filled with low probability adversarial pockets. Similarly, b) \citeauthor{Tanay2016}'s hypothesis indicates the existence of tilted boundaries what contributes to the emergence of adversarial examples.}
    \label{fig:tilting}
    \vspace{-5mm}
\end{figure}

\citeauthor{Tanay2016} \cite{Tanay2016}, on the other hand, have rejected the linear hypothesis proposed by \citeauthor{Goodfellow2014} by assuming that it is "insufficient" and "unconvincing". They have proposed instead a \textit{boundary tilting perspective} to explain the adversarial paradox. This assumption, according to the authors, is more related to the explanation given by \citeauthor{Szegedy2013}, where a learnt class boundary lies close to the training samples manifold, but this learnt boundary is "tilted" with respect to this training manifold. Thereby, adversarial images can be generated by perturbing legitimate samples towards the classification boundary until they cross it. The amount of required perturbation is smaller as the tilting degree decreases, producing high-confidence and misleading adversarial examples, containing visually imperceptible perturbations. The authors also believe this effect might be a result of an overfitted model. Figure \ref{fig:tilting}b shows a simplified illustration of the boundary tilting perspective compared with the \citeauthor{Szegedy2013} hypothesis.

\subsection{High Dimensional Manifold}

\citeauthor{gilmer2018adversarial} \cite{gilmer2018adversarial}, in concordance with other works such as \citeauthor{shafahi2018adversarial, mahloujifar2019curse} and \citeauthor{fawzi2018adversarial} \cite{shafahi2018adversarial, mahloujifar2019curse, fawzi2018adversarial}, said that the phenomenon of adversarial examples is result from the high dimensional nature of the data manifold. In order to show evidences, \citeauthor{gilmer2018adversarial} have created a synthetic dataset for better controlling their experiments, and used it afterwards to train a model. After training it, the authors observed that inputs correctly classified by the model were close to nearby misclassified adversarial inputs, meaning that learning models are necessarily vulnerable to adversarial examples, independently of the training procedure used. At last, based on empirical results, \citeauthor{gilmer2018adversarial} have also denied the assumption that states adversarial examples lie on a different distribution when compared to legitimate data \cite{Meng2017, song2017pixeldefend, samangouei2018defense, ghosh2019resisting}.

\subsection{Lack of Enough Training Data}
\citeauthor{schmidt2018adversarially} \cite{schmidt2018adversarially} claim learning models must generalize in a strong sense, \textit{i.e.} with the help of robust optimization, in order to achieve robustness. Basically, the authors observed the existence of adversarial examples is not necessarily a shortcoming of specific classification models, but an unavoidable consequence of working in a statistical setting. After gathering some empirical results, the authors concluded that, currently, there are no working approaches which attain adversarial robustness mainly because existing datasets are not large enough to train strong classifiers.

\subsection{Non-Robust Features Hypothesis}

\citeauthor{ilyas2019adversarial} \cite{ilyas2019adversarial} have provided a different explanation based on the assumption the existence of adversarial perturbations does not necessarily indicate flaws regarding learning models or training procedures, but actually regarding images' features. By taking into account the human's perception, the authors split the features into (i) robust features, that lead models to correctly predict the true class even when they are adversarially perturbed, and (ii) non-robust features, which are features derived from patterns in the data distribution that
are highly predictive yet brittle, incomprehensible to humans and more susceptible to be perturbed by an adversary. In order to underlie their assumption, the authors proposed constructing a novel dataset formed by images containing solely robust features which have been filtered out from the original input images by using the logits layer of a trained DNN. Then, this dataset has been used to train another DNN that has been used to perform a comparative study. The results has led the authors to find evidences that adversarial examples might really arise as a result of the presence of non-robust features, what goes in a opposite direction of what it is commonly believed: that adversarial examples are not necessarily tied to the standard training framework. Their conclusion is somehow related to the \citeauthor{schmidt2018adversarially} \cite{schmidt2018adversarially} work.
\vspace{-2mm}

\subsection{Explanations for Adversarial Transferability} \label{subsec:transferability}

As briefly mentioned in Section \ref{subsub:adv_training}, adversarial examples make heavy use of the transferability property to drastically affect the performance of learning models even in more realistic scenarios, where the attacker does not have access to much or any information regarding the target classifier, as is simulated by grey and black-box settings. 
Adversarial transferability can be formally defined as the property that some adversarial samples have to mislead not only a target model $f$, but also other models $f'$ even when their architectures greatly differ \cite{Papernot2016c}. \citeauthor{Papernot2016d} \cite{Papernot2016d} have split adversarial transferability into two main categories: (i) \textit{intra-technique transferability}, which occurs between two models that share a similar learning algorithm (e.g. DNNs) and are trained using the same dataset, however initialized with different parameters (e.g. transferability between two DNN architectures, such as VGG-16 and ResNet-152); (ii) \textit{cross-technique transferability}, that occurs between two models which respectivelly belong to different learning algorithms (e.g. DNN and SVM), where can even perform different learning tasks, such as image classification and object detection (see Appendix \ref{subsec:other_tasks} for more details). According to \citeauthor{wiyatno2019adversarial} \cite{wiyatno2019adversarial}, understanding the transferability phenomenon is critical not only to explain the existence of adversarial examples, but to create safer machine learning
models.

Some assumptions have arisen in literature as an attempt to explain adversarial transferability. The linearity hypothesis assumed by \citeauthor{Goodfellow2014} \cite{Goodfellow2014} suggests the direction of perturbations may be the crucial factor that allows the adversarial effect transfer among models, since the disturbances end up acquiring similar functions through training. \citeauthor{Tramer2017} \cite{Tramer2017}, in turn, have hypothesized if adversarial transferability is actually a consequence of the intersection between the adversarial subspace of two different models. By estimating the number of orthogonal adversarial directions using a techinique called \textit{Gradient Aligned Adversarial Subspace (GAAS)}, they found that the separating distance between the decision boundaries of two models was, on average, smaller than the distances between any inputs to the decision boundaries, even on
adversarially trained models. This suggests their adversarial subspaces were overlapped. At last, they also concluded that transferability is inherent to models that preserve non-robust properties when trying to learn feature representations of the input space, what according to the authors is not a consequence of a lack of robustness, but actually a intrinsic property of the learning algorithms themselves. Their findings agreed with the works of \citeauthor{Liu2016} \cite{Liu2016} and \citeauthor{ilyas2019adversarial} \cite{ilyas2019adversarial}.
\vspace{-2mm}

%% file: sections/sec6.tex
{Defending robustly against adversarial attacks is still an open question. \citeauthor{carlini2019evaluating} \cite{carlini2019evaluating} assert defenses often claim robustness against adversarial examples without carrying out common security evaluations}, what in fact contributes to the construction of brittle and limited architectures which are rapidly broken by novel and adaptive attacks. For this reason, the authors have defined a basic set of principles and methodologies that should be followed by both defenders and reviewers to check whether a defense evaluation is thorough and follows currently accepted best practices. This is crucial to prevent researchers from taking deceitful statements and conclusions about their works. In the following are listed and briefly explained some basic and relevant principles based on the \citeauthor{carlini2019evaluating}'s guide for properly evaluating general defenses. For further orientations, it is recommended consult the authors' paper \cite{carlini2019evaluating}.

\subsection{Define a Threat Model}

A defense should always define a threat model where it states to be robust against adversarial attacks. It is important the threat model be described in details, preferably following the taxonomy defined in Section \ref{sec:tax_threat}, so that reviewers and attackers can restrict their evaluations under the requirements the defense affirms to be secure. For instance, a certain defense claims robustness under a threat model formed by: evasive attacks conducted in a white-box scenario where adversarial examples are generated by gradient and approximation-based attacks with $L_2$ norm and perturbation size less than 1.5. Based on this information, fair attackers which are interested in this defense must follow exactly what is specified by this threat model when designing their attacks.

\subsection{Simulate Adaptive Adversaries}

A good evaluation must test the limits of a defense by simulating adaptive adversaries which make use of its threat model to elaborate strong attacks. All settings and attacks scenarios that stand a chance to bypass the defense should be taken into consideration without exceptions. An evaluation conducted only based on non-adaptive adversaries is of very limited utility since the results produced by the experiments do not bring reliable conclusions that support the defense's claims and its robustness bounds. A good evaluation will not try to support or assist the defense's claims, but will try to break it under its threat model at all costs. Therefore, weak attack settings and algorithms, such as FGSM attack\footnote{FGSM originally was implemented to support the linearity hypothesis made in \citeauthor{Goodfellow2014} \cite{Goodfellow2014}. For this and other reasons related to the attack configurations, such as its sequential execution when computing perturbations, this attack is considered weak and untrustworthy to fully test defenses, usually used only to run sanity tests (see Section \ref{sec:sanity_tests}).} must not be solely used. It is worth mentioning there are some relevant libraries available online for helping researchers to perform evaluations by simulating adaptive adversaries, such as \textit{CleverHans} \cite{papernot2018cleverhans}, \textit{Adversarial Robustness Toolbox (ART)} \cite{art2018}, \textit{Foolbox} \cite{rauber2017foolbox}, \textit{DEEPSEC \cite{ling2019deepsec}} and \textit{AdvBox} \cite{goodman2020advbox}.

\subsection{Develop Provable Lower Bounds of Robustness}\label{sec:lower_bounds}

Most works make use of empirical and heuristic evaluations to assess the robustness of their defenses. However, provable approaches are preferred since they provide, when the proof is correct, lower bounds of robustness which ensure the performance of the evaluated defense will never fall below that level. Nevertheless, provable evaluations usually suffer from the lack of generalization, since they get attached to the network architecture and a specific set of adversarial examples $\mathcal{X}$, crafted using a certain attack algorithm, that has been used in the experiments. This evaluation does not give any proofs that extend for other adversarial examples $x' \notin \mathcal{X}$, what makes the statement less powerful. Circumvent these problems when developing provable lower bounds is an active research path (see Section \ref{subsec:lower_robustness}).

\subsection{Perform Basic Sanity Tests} \label{sec:sanity_tests}

Sanity tests are important to identify anomalies and antagonic results that can lead authors to take incorrect conclusions. \citeauthor{carlini2019evaluating} \cite{carlini2019evaluating} has listed some basic sanity tests that should be run to complement the experiments and support the results.

\begin{itemize}[leftmargin=*]
    \item \textbf{Report model accuracy on legitimate samples:} {while the protection of learning models against adversarial examples is a relevant security issue}, a significant decrease on legitimate data on behalf of increasing the robustness of the model might be unreasonable for scenarios where the probability of an actual adversarial attack is low and the cost of a misclassification is not high. For reactive defenses, it is important to evaluate how the rejection of perturbed samples can affect the accuracy of the model on legitimate samples. An analysis of a Receiver Operating Characteristic (ROC) curve may be helpful to check how the choice of a threshold for rejecting adversarial inputs can decrease the model's clean accuracy;
    
    \item \textbf{Iterative \textit{vs.} sequential attacks:} iterative attacks are more powerful than sequential attacks. If adversarial examples crafted by a sequential algorithm are able to affect classification models more than examples crafted by iterative ones, it can indicate that the iterative attack is not properly calibrated;
    
    \item \textbf{Increase the perturbation budget:} attacks when allowed to produce larger amounts of distortion in images usually fool classifiers more often than attacks with smaller perturbation budgets. Therefore, if the attack success rate decreases as the perturbation budget increases, this attack algorithm is likely flawed;
    
    \item \textbf{Try brute force attacks:} it can be an alternative in scenarios where the attacks do not succeed very often. By performing a random search attack within the defense's threat model can help the attacker or reviewer to find adversarial examples which have not been found by standard adversarial attacks, what indicates these algorithms must be somehow improved. \citeauthor{carlini2019evaluating} recommend starting this sanity test by sampling random points at larger distances from the legitimate input, limiting the search to strictly smaller distortions whenever an adversarial example is found.
    
    \item \textbf{White-box vs. black-box attacks:} white-box attacks are generally more powerful than black-box attacks, since the attacker has complete access to the model and its parameters. For this reason, gradient-based attacks should, in principle, present better success rates. If gradient-based attacks have worse performance when compared to other attack approaches, it can indicate the defense is somehow performing a kind of gradient masking and the gradient-based attack needs calibration. 
    
    \item \textbf{Attack similar undefended models:} proactive and reactive defenses typically introduce a couple of modifications in the networks in order to increase their robustness. However, it can be worth trying remove these security components out from the model and evaluate it under attacks without any protection. If the undefended model appears robust nevertheless, it can infer that the defense itself is not  actually protecting the model.  
\end{itemize}

\subsection{Releasing of Source Code}

It is crucial that all source code used to implement the experiments and pre-trained models referred in the defense's paper, including even their hyperparameters, be available to the community through online repositories so that interested reviewers can reproduce the evaluations made by the original work and ensure their correctness. 



%% file: sections/sec7.tex
During the development of this paper, it has been noticed adversarial defenses are still in their infancy, despite the impressive growth of published works in the last years. There are numerous important questions waiting for answers, specially those referring to how defend robustly against adversarial examples. This opens some promising research paths that will be detailed in the following.

\subsection{Development of Theoretical Lower Bounds of Robustness}\label{subsec:lower_robustness}

Most defenses are limited to empirical evaluations and do not claim robustness to unknown attacks \cite{wiyatno2019adversarial}. Other works, in turn, devise theoretical robustness bounds which do not generalize to different attacks and threat models studied. A promising research path is the investigation of properties that can theoretically guarantee general lower bounds of robustness to adversarial attacks (see Section \ref{sec:lower_bounds}).

\subsection{Unanimously Explanations Concerning the Existence and Transferability of Adversarial Examples}

As can be seen in Section \ref{sec:5}, there are already some explanations for the existence and transferability of adversarial examples. However, none of them is universally accepted due to the lack of proofs. Developing unanimously provable explanations for these phenomena is relevant for the field of Adversarial Machine Learning, since they will guide future defenses to focus on solving the actual flaw and help the community understand better the inner workings of deep learning models.

\subsection{Devising of Efficient Attack Algorithms}

Crafting strong adversarial examples is computationally expensive even on vanilla datasets. Applications which count on small response times, such as the traffic signal recognition system of a autonomous vehicle, require efficiency from attack algorithms when intercepting and perturbing the inputs. Attacks in black-box environments also impose more difficulties to the attacker since he usually has available only a limited number of queries to the oracle\footnote{In black-box attacks, the term oracle often represents the target model the attacker wants to fool.} in order to generate the perturbations. Regarding defenses, a good evaluation also needs testing numerous attacks, what can be computationally infeasible depending on the algorithms and datasets used. Therefore, devising strong and efficient adversarial attacks is a relevant research path for both fields of Adversarial Machine Learning.

\subsection{Comparison to Prior Work}

As previously mentioned, a large amount of adversarial defenses emerges in literature, however few of them perform a comparative study with other methods. A well-conducted study comparing different security approaches could help fomenting results in addition to reveal promising architectures for specific threat models.

\subsection{Development of Hybrid Defense Architectures}

It is worth mentioning as a encouraging research path the development of hybrid defenses. The term \textit{hybrid defense} stands for an architecture formed by different countermeasures which are organized on individual processes, called \textit{modules}. Each module would be responsable for performing some security procedure according to the approach that represents it. On each module, a component (represented by a defense or preprocessing method) would be randomly picked from a repository when receiving the input image. For instance, a hybrid defense could consist of three modules: a \textit{(i) reactive module}, which randomly chooses a reactive defense from a repository to detect adversarial images; a \textit{(ii) preprocessing module} that randomly processes the detected adversarial images that came from the reactive module, and a \textit{(iii) proactive module} which similarly chooses at random a proactive defense to finally classify the input image. To the best of knowledge, it has not been found in literature any work that has adopted similar approach or study, what, in turn, make this path open for future opportunities.

%% file: sections/sec8.tex
Deep Learning models have revealed to be susceptible to attacks of adversarial nature despite shown impressive abilities when solving complex cognitive problems, specially tasks related to Computer Vision, such as image classification and recognition. This vulnerability severely menaces the application of these learning algorithms in safety-critical scenarios, what in turn may jeopardize the development of the field if this security issue persists in the future. The scientific community has been struggling to find alternatives to defend against adversarial attacks practically since this problem was firstly spotted by the work of \citeauthor{Szegedy2013} \cite{Szegedy2013}. The numerous proposed defenses, albeit promising at first, have shown to be brittle and innefective to stop strong and adaptive attacks though. This arms race between attacks and defenses makes the field of Adversarial Machine Learning fairly dynamic and active, where the emergence of novel defense approaches almost daily plays a role in becoming review papers quickly outdated. 

Before this chaotic scenario, this work has aimed to attend the interested readerships by elaborating a comprehensive and self-contained survey that gathers the most relevant research on Adversarial Machine Learning. It has covered topics regarding since Machine Learning basics to adversarial examples and attacks, nevertheless with an emphasis on giving the readers a defender's perspective. An extensive review the literature has allowed recent and promising defenses, not yet mentioned by other works, be studied and categorized following a novel taxonomy. Moreover, existing taxonomies to organize adversarial examples and attacks have been updated in order to cover further approaches. Furthermore, it has been gathered existing relevent explanations for the existence and transferability of adversarial examples, listed some common policies that should be considered by both defenders and reviewers when respectivelly designing and evaluating security methods for deep learning models and provided some promising paths for future work. In summary, the main contributions of this work were the following:

\begin{itemize}[leftmargin=*]
    \item The provision of a background regarding CNNs and some relevant architectures present in literature ranked according to their respective performance in the ILSVRC top-5 classification challenge from 2012 to 2017. It was also highlighted other important Deep Learning algorithms in Adversarial Machine Learning, such as Autoencoders and Generative Adversarial Networks (GANs); \vspace{3mm}
    
    \item The update of some existing taxonomies to categorize different types of adversarial images and novel attack approaches that have raised in literature;\vspace{3mm}
    
    \item A exhaustive review and discussion of defenses against adversarial attacks that were categorized using a novel taxonomy;\vspace{3mm}
    
    \item The address of relevant explanations for the existence and transferability of adversarial examples;\vspace{3mm}
    
    \item The discussion of promising research paths for future works on Adversarial Machine Learning.
\end{itemize}

Securing against adversarial attacks is crucial for the future of several applications. Therefore, this paper has been elaborated to provide a detailed overview of the area in order to help researchers to devise better and stronger defenses. For the best of the authors' knowledge, this is the most comprehensive survey focused on adversarial defenses available in literature and it is hoped that this work can help the community to make Deep Learning models reaching their prime-time soon. 





%% file: sections/appendix.tex
{\section{Other Tasks in Adversarial Machine Learning for Computer Vision}} \label{subsec:other_tasks}

Besides Image Classification, Adversarial Machine Learning also takes part in various other tasks of Computer Vision. Among the mainstream options, two tasks in specific are widely approached in papers: \textit{(i) {Object Detection}} and \textit{(ii) {Semantic Segmentation}}. Object Detection tasks aim to identify semantic objects in input images by surrounding each of them usually by drawing a rectangle, also known as \textit{bounding box}, around the detected objects. In turn, Semantic Segmentation aims to represent an image into something more meaningful and easier to analyze by assigning a label for each pixel in the input image which shares similar characteristics \cite{garcia2017review}. Table \ref{tbl:appendixTasks} references for interested readers some related works in Adversarial Machine Learning towards Object Detection and Semantic Segmentation.
\vspace{3mm}

\begin{table}[h!]
\caption{Relevant works on Adversarial Machine Learning for Object Detection and Image Segmentation tasks.}
\resizebox{5cm}{!}{%
\begin{threeparttable}
\begin{tabular}{@{}lc@{}}
\toprule
Work and Reference & Task\tnote{*} \\ \midrule
\citeauthor{xie2017dag} \cite{xie2017dag} & OBJ, SGS \\
\citeauthor{metzen2017universal} \cite{metzen2017universal} & SGS \\
\citeauthor{moosavi2017universal} \cite{moosavi2017universal} & SGS \\
\citeauthor{fischer2017adversarial} \cite{fischer2017adversarial} & SGS \\ \citeauthor{lu2017no} \cite{lu2017no} & OBJ \\ \citeauthor{chen2018shapeshifter} \cite{chen2018shapeshifter} & OBJ \\ \citeauthor{thys2019fooling} \cite{thys2019fooling} & OBJ \\
\bottomrule
\end{tabular}
\begin{tablenotes}[flushleft]
  \footnotesize
  \item[*]OBJ: \textit{Object Detection}; SGS: \textit{Semantic Segmentation}.
\end{tablenotes}
\end{threeparttable}
\label{tbl:appendixTasks}}
\end{table}

\section{Standard Datasets in Computer Vision}

Datasets are important tools for evaluating Deep Learning algorithms. In the field of Adversarial Machine Learning and Computer Vision, some most used datasets are summarized by Table \ref{tbl:datasets}.

\begin{table}[h!]
\caption{Popular Datasets in Adversarial Machine Learning for Computer Vision}
\begin{threeparttable}
\resizebox{\textwidth}{!}{%
\begin{tabular}{@{}lcccccccc@{}}
\toprule
{Name and Reference} & Main Task\tnote{*} & Year & Classes & Images' Resolution & Training Samples & Validation Samples & Testing Samples & Total of Images \\ \midrule
MNIST \cite{lecun1998mnist} & ICR & 1998 & 10 & 28x28x1 & 60,000 & N/A & 10,000 & 70,000 \\
CIFAR-10 \cite{cifar10} & ICR & 2009 & 10 & 32x32x3 & 50,000 & N/A & 10,000 & 60,000 \\
CIFAR-100 \cite{cifar10} & ICR & 2009 & 100 & 32x32x3 & 50,000 & N/A & 10,000 & 60,000 \\
SVHN \cite{svhn} & ICR / OBJ & 2011 & 10 & 32x32x3 & 73,257 & 531,131 & 26,032 & 630,420 \\
GTSRB \cite{gtsrb} & ICR / OBJ & 2012 & 43 & {[}15x15x3, 250x250x3{]} & 34,799 & 4,410 & 12,630 & 51,839 \\
ImageNet \cite{imagenetRussakovsky} & ICR / OBJ & 2015 & 1,000 & 482x415x3 (average) & N/A & N/A & N/A & 14,197,122 \\
CelebA \cite{celeba} & ICR / OBJ & 2015 & 10,177 & 218x178x3 & 162,770 & 19,867 & 19,962 & 202,599 \\
VOC2012 \cite{voc2012} & ICR / OBJ / SGS & 2012 & 20 & 469x387x3 (average) & 5,717 & 5,823 & 10,991 & 22,531 \\
MS-COCO \cite{mscoco} & OBJ / SGS & 2014 & 171 & 640x480x3 & 165,482 & 81,208 & 81,434 & 328,124 \\
STL-10 \cite{stl10} & ICR & 2011 & 10 & 96x96x3 & 5,000 & 100,000 (unlabeled) & 8,000 & 113,000 \\
Toronto Faces Dataset \cite{toronto_dataset} & ICR / OBJ & 2010 & 7 & 32x32x3 & 2,925 & 98,058 (unlabeled) & 418 & 101,401 \\ \bottomrule
\end{tabular}}
\begin{tablenotes}[flushleft]
  \footnotesize
  \item[*]ICR: \textit{Image Classification and Recognition}; OBJ: \textit{Object Detection}; SGS: \textit{Semantic Segmentation}; N/A: \textit{Not Available}.
\end{tablenotes}
\end{threeparttable}
\label{tbl:datasets}
\end{table}